\documentclass[10pt,twocolumn,letterpaper]{article}

\usepackage{cvpr}            

\definecolor{cvprblue}{rgb}{0.21,0.49,0.74}
\usepackage[pagebackref,breaklinks,colorlinks,allcolors=cvprblue]{hyperref}

\usepackage{hyperref}
\usepackage{url}
\usepackage{booktabs}
\usepackage{graphicx}
\usepackage{amsmath}
\usepackage{multirow}
\usepackage{wrapfig}
\usepackage{overpic}
\usepackage{float}
\usepackage{tablefootnote}
\usepackage{tabularx}
\usepackage[table]{xcolor}
\usepackage{amssymb}

\title{WeMMU: Enhanced Bridging of Vision-Language Models \\ and Diffusion Models via Noisy Query Tokens}

\author{%
  \textbf{Jian Yang}\textsuperscript{1,}~\footnotemark[1]%
  \quad%
  \textbf{Dacheng Yin}~\footnotemark[1]~\textsuperscript{, }\footnotemark[2]%
  \quad%
  \textbf{Xiaoxuan He}\textsuperscript{2}%
  \quad%
  \textbf{Yong Li}\textsuperscript{3}%
  \quad%
  \textbf{Fengyun Rao}
  \\
  \quad%
  \textbf{Jing LYU}
  \quad%
  \textbf{Wei Zhai}\textsuperscript{1}%
  \quad%
  \textbf{Yang Cao}\textsuperscript{1},%
  \quad%
  \textbf{Zheng-Jun Zha}\textsuperscript{1,}\footnotemark[3] \\
  \textsuperscript{1} MoE Key Laboratory of Brain-inspired Intelligent Perception and Cognition, \\
  \hspace*{1.5em}University of Science and Technology of China \\
  \textsuperscript{2} ZheJiang University~~~~
  \textsuperscript{3} The Hong Kong University of Science and Technology \\
  \small%
  \texttt{\{yangjian12138@mail., wzhai056@, forrest@, zhazj@\}ustc.edu.cn}\\
  \small%
  \texttt{xiaoxuanhe@zju.edu.cn}~~~~
  \small%
  \texttt{yong.li@connect.ust.hk}
  \quad%
}

\begin{document}
\maketitle
\renewcommand{\thefootnote}{\fnsymbol{footnote}}
\footnotetext[1]{Co-first Author}
\footnotetext[2]{Project Leader}
\footnotetext[3]{Corresponding Author}
\renewcommand{\thefootnote}{\arabic{footnote}} 

\begin{abstract}
Recent progress in multimodal large language models (MLLMs) has highlighted the challenge of efficiently bridging pre-trained Vision-Language Models (VLMs) with Diffusion Models. While methods using a fixed number of learnable query tokens offer computational efficiency, they suffer from task generalization collapse, failing to adapt to new tasks that are distant from their pre-training tasks. To overcome this, we propose Noisy Query Tokens, which learn a distributed representation space between the VLM and Diffusion Model via end-to-end optimization, enhancing continual learning. Additionally, we introduce a VAE branch with linear projection to recover fine-grained image details. Experimental results confirm our approach mitigates generalization collapse and enables stable continual learning across diverse tasks.
\end{abstract}    
\section{Introduction}
\label{sec:intro}

In recent years, significant progress in multimodal understanding and generation has prompted researchers to integrate these two capabilities into a single model, enhancing interaction diversity. Consequently, the development of unified multimodal large language models (MLLMs) has garnered substantial attention.
However, early efforts, such as Chameleon~\citep{chameleon} and Transfusion~\citep{transfusion}, highlighted the difficulty of adapting models to handle both tasks simultaneously. This often resulted in a performance trade-off during training, as the two tasks competed for the model's resources.
Subsequently, the Janus-flow~\citep{janusflow} approach addressed this challenge by introducing a specialized generation encoder, enabling both understanding and generation to coexist effectively within a unified framework.

Building on these insights, research has gradually developed along two primary directions:
(1) The first integrates insights from Janus-flow~\citep{janusflow}, focusing on training diffusion-based generation parameters directly within a unified MLLM architecture. Models like Mogao~\citep{mogao} and Bagel~\citep{bagel} employ specialized generation pathways to achieve deep semantic alignment.
(2)  The second explores how to connect pre-trained Vision-Language Models (VLMs) and Diffusion Models. The goal is to achieve significant performance with high efficiency, while maintaining the flexibility to combine various pre-trained models. The MetaQueries~\citep{metaquery} series, which employs learnable query tokens to establish this connection, stands as a representative example of this approach.

\begin{figure}
    \centering
    \includegraphics[width=0.90\linewidth]{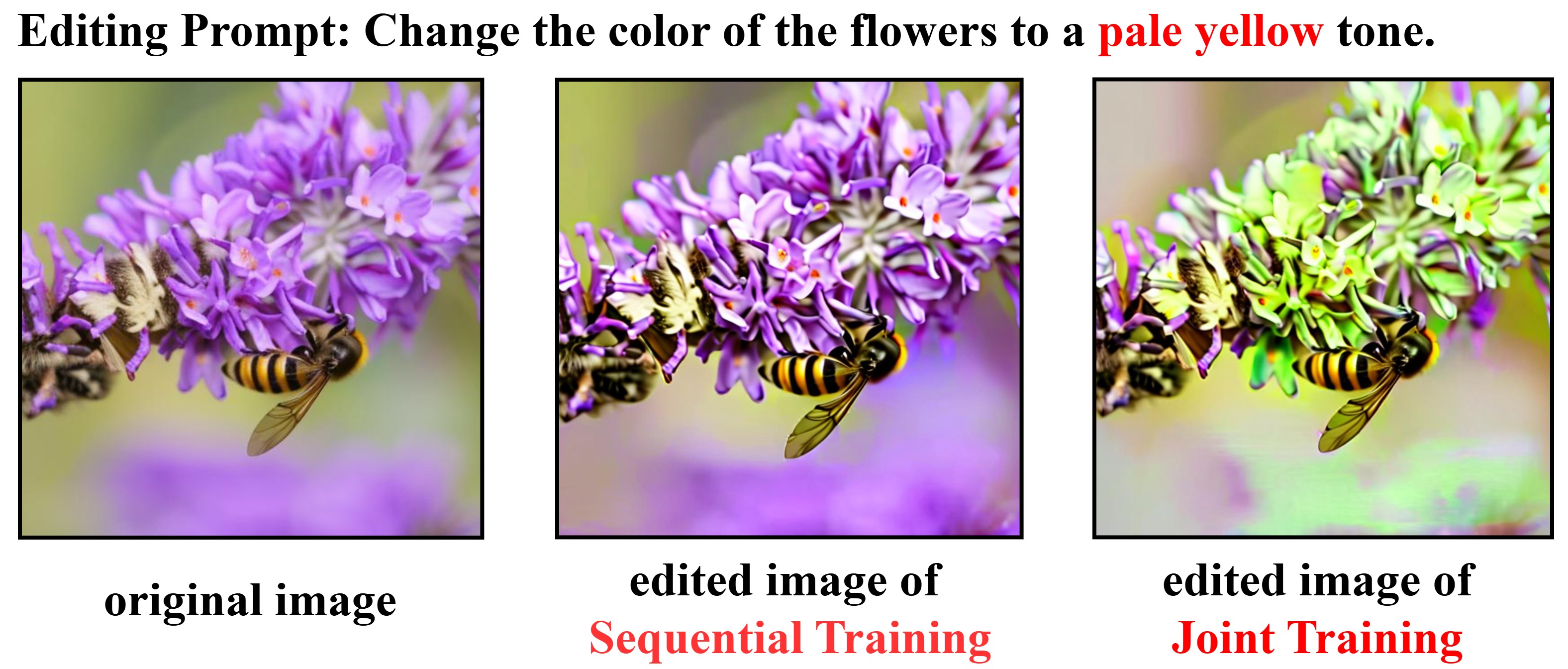}
    \caption{Task Generalization Collapse. Sequential training (middle) fails editing, merely reconstructs the input. Joint training (right) works but is unsustainable, requiring full retraining for new tasks.}
    \label{fig:problem}
    \vspace{-1.0em}
\end{figure}

Our work aims to develop a highly efficient and flexible Unified MLLM framework. Achieving this involves navigating a fundamental trade-off between two primary approaches. Although powerful, the former approach requires training image generation from scratch, which demands massive data and computational resources, thus raising the training threshold. In contrast, the second approach demonstrates exceptional efficiency, capable of aligning a VLM and a Diffusion Model with only 25M data instances (4 epochs) to achieve competitive performance~\citep{metaquery}. This high efficiency makes the second approach highly promising for scenarios with limited data and computational resources. However, we found that the latter approach, using learnable query tokens, encounters severe task generalization collapse. Specifically, after pre-training on ``text-to-image generation'' and ``image reconstruction'' tasks~\citep{recA}, the learnable query tokens tend to become ``rigid''. For instance, when performing a subsequent ``image editing'' task, the model ignores the textual instruction and tends to mechanically reconstruct the input image. This indicates that the model fails to generalize to new tasks that significantly differ from the pre-training tasks, a phenomenon we refer to as ``task generalization collapse''.
Consequently, the methods with learnable query tokens must return to early training stages and simultaneously retrain on all task types when tackling new and significantly different tasks, as illustrated in Fig.~\ref{fig:problem}. \textbf{However, this approach is not conducive to the model's sustainable learning and application expansion.}
To tackle the task generalization collapse of learnable query tokens, a straightforward idea is to fix the pre-trained models and only fine-tune or re-initialize the learnable query tokens for new tasks. However, we observed that this leads to rapid training collapse.
We hypothesize that this collapse occurs because the learnable query tokens quickly converge to a fixed, task-specific ``mean point" with limited expressive power, preventing generalization to diverse new tasks. This inspired our proposal: the intermediate representation bridging the VLM and the generative model should not be an isolated point, but a distribution.

To this end, we propose a novel framework that reformulates the connector as a Probabilistic Expert Bridge. First, to tackle task generalization collapse, we introduce Noisy Query Tokens sampled from a distribution $(\mathcal{N}(0, \textit{I}))$ rather than using deterministic embeddings. This randomness prevents the model from relying on task-specific shortcuts (overfitting to reconstruction) and compels the VLM to learn robust, generalized mappings.
Second, to effectively project these noisy tokens into valid generation conditions, we adapt the Expert Pathway architecture~\citep{bagel}. Unlike Bagel, which uses this pathway to learn generation from scratch, we leverage it solely to align the VLM's outputs with a pre-trained Diffusion Model. This design synthesizes the strengths of both paradigms: it retains the deep semantic alignment capability of the Expert Pathway and the robustness of Noisy Queries, while preserving the high efficiency of using a pre-trained generative backbone.

Beyond task generalization, preserving fine-grained image details remains a challenge, as VLMs often lose high-frequency information during processing.
A common remedy is to inject supplementary visual information (e.g., VAE features) directly into the Diffusion Model. 
However, we argue that this approach creates a structural redundancy. It forces the generative backbone to take on the additional role of multimodal context aggregation—handling complex, interleaved inputs (e.g., multiple reference images)—a task that partially overlaps with the MLLM's capabilities.
To achieve a more efficient ``division of labor'', we propose injecting VAE features directly into the VLM via a lightweight projection. This allows the VLM to act as a comprehensive processor, fusing text instructions, high-level semantics, and low-level details into the Noisy Query Tokens. Consequently, the Diffusion Model is relieved of the burden of feature alignment and can focus exclusively on its core competency: high-fidelity denoising and generation.

In summary, we tackle the task generalization collapse of learnable query tokens by introducing Noisy Query Tokens and enhance image fidelity by incorporating a VAE branch.
Our main contributions are as follows:
\begin{itemize}
\item We identify and analyze the task generalization collapse problem in methods using learnable query tokens, highlighting the need for sustainable learning in unified MLLMs.
\item We introduce Noisy Query Tokens within a Probabilistic Expert Bridge. This approach solves the collapse issue by enforcing distribution learning and achieves deep semantic alignment without the heavy cost of training generation from scratch.
\item We propose a Feature Integration Strategy (Division of Labor) that injects visual details into the MLLM rather than the diffusion model, maximizing the architectural strengths of both components.
\end{itemize}
\section{Related Works}
\label{sec:related}

\begin{figure*}
    \centering
   \vspace{-1em}
    \includegraphics[width=0.95\linewidth]{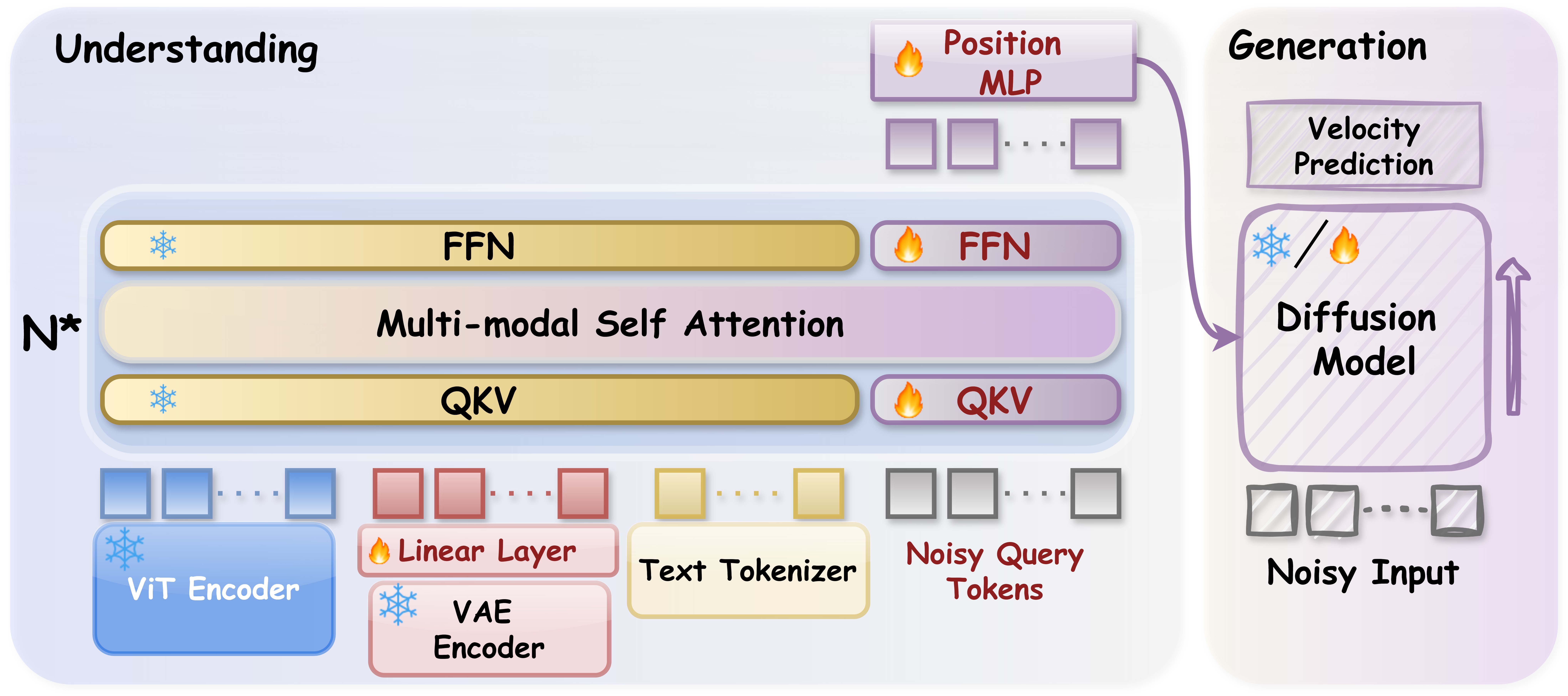}
    \caption{Overall framework. We bridge a frozen VLM and tunable Diffusion Model via Noisy Query Tokens, sampled from $\mathcal{N}(0,\textit{I})$ per step. These tokens aggregate image and text features in a parallel generation pathway, while the original VLM pathway remains frozen. A VAE branch injects fine-grained details via a linear layer. Position MLP adds 2D spatial cues and projects features to condition the diffusion model. This design maintains clear labor division: VLM handles understanding, diffusion model focuses on generation.}
    \label{fig:method}
    \vspace{-1em}
\end{figure*}

\subsection{Unified Architectures: Integrating Generation into MLLMs}
This line of research aims to build an end-to-end Multimodal Large Language Model (MLLM) with native image generation capabilities. While promising, this approach typically requires large-scale datasets and substantial computational resources for training or fine-tuning.

\subsubsection{LLM as Diffusion models}
Some studies explore deeply integrating diffusion concepts into Transformers. Works like Transfusion~\citep{transfusion}, Janus-Flow~\citep{janusflow}, and LMFusion~\citep{lmfusion} attempt to replace the U-Net~\citep{u-net} in traditional diffusion models with Transformer blocks or directly employ Large Language Models (LLMs) as the denoiser. Models such as Bagel~\citep{bagel} and Mogao~\citep{mogao}, trained on massive image-text data, aim to equip the model with native instruction-following abilities for high-quality generation. Although these models achieve impressive results in fidelity and instruction adherence, their prohibitive training costs limit widespread adoption and rapid iteration.

\subsubsection{Hybrid Paradigms and Alternative Strategies}
Another line of research strives to co-locate understanding and generation within a unified framework, but adopts more diverse strategies. Early models like MMAR~\citep{mmar}, Chameleon~\citep{chameleon}, and EMU-3~\citep{emu3} shared most parameters, often causing performance interference. To mitigate this, models like Janus~\citep{janus} and Janus-Pro~\citep{janus-pro}  introduced a dedicated generation encoder, establishing a ``division of labor'' design. Meanwhile, auto-regressive models like VARGPT~\citep{vargpt} and OneCat~\citep{onecat} serialize image tokens for LLM prediction, a paradigm that is generally less efficient than the parallel denoising in diffusion models.

\subsection{Bridging Architectures: Connecting Understanding and Generation Modules}

Given the availability of powerful pre-trained models for understanding (e.g., VLMs like QWen2.5VL~\citep{qwen2p5VL}) and generation (e.g., SD3~\citep{sd3}, Sana 1.5~\citep{sana1p5}, FLUX-dev~\citep{flux-dev}), a highly efficient alternative is to connect them with a lightweight ``bridge", minimizing trainable costs. Based on the granularity of information transferred by the bridging module, these approaches can be further categorized.

One line of work, exemplified by models like Ovis-U1~\citep{ovis-u1}, OmniGen2~\citep{omnigen2}, Query-Kontext~\citep{querykontext}, and Uniworld-V1~\citep{uniworld} conditions the diffusion model on \textbf{dense feature embeddings} from the VLM. This provides rich context but burdens the generator with filtering complex signals, increasing training difficulty.

Another line argues for a clearer role division: the VLM should interpret instructions and compress key generation conditions (e.g., content, style) into a compact representation for the diffusion model. Representative methods include MetaQueries~\citep{metaquery}, TBAC-UniImage~\citep{tbac}, which employ a set of learnable query tokens as the bridge. Bifrost-1~\citep{bifrost} uses a similar alignment module. This strategy reduces the burden on the diffusion model.

However, a critical limitation of methods relying on a fixed set of learnable query tokens is task generalization collapse. These tokens tend to overfit to the average representations of training tasks, leading to significant performance drops on new, dissimilar tasks. Our work is directly motivated by the need to overcome this fundamental limitation.

\section{Method}
\label{sec:method}

\subsection{Preliminaries}
Our method utilizes the flow matching objective~\citep{flowmatching} for end-to-end parameter optimization. Therefore, this section briefly introduces flow matching~\citep{flowmatching} and its variants used in our approach.

\paragraph{Flow-Matching and Conditional Flow Matching.} Flow Matching (FM)~\citep{flowmatching} is a framework for training continuous-time generative models by learning a vector field that defines a probability path from a simple prior distribution (e.g., a Gaussian) to a complex data distribution.
This evolution is described by an ordinary differential equation (ODE):
\begin{equation}
    \frac{d{x_t}}{dt}=v(x_t, t)  
\end{equation}
where $x_t$ is the state at time $t \in [0, 1]$ , and $v(x_t, t)$ is the time-dependent vector field to be learned.

To simplify training, Conditional Flow Matching (CFM)~\citep{flowmatching} is often used. Instead of learning the complex vector field of the entire data distribution, we learn how to move a single noise point $\epsilon \sim \mathcal{N}(0,\textit{I})$ to a single data point $x_0$. The simplest way to do this is to define a straight path between them:
\begin{equation}
    x_t=(1-t)x_0 + t\epsilon
\end{equation}
The vector field corresponding to this path is simply $\epsilon - x_0$.
The training objective, therefore, is to make a neural network $v_{\theta}(x_t, t, y)$ predict this vector, where $y$ is the condition vector. The loss function is:
\begin{equation}
    \mathcal{L}_{CFM}(\theta) = \mathbb{E}[\left\| v_{\theta}(x_t, t, y) - (\epsilon - x_0) \right\|^2]
    \label{ConditionFM}
\end{equation}

\paragraph{Contrastive Flow-Matching.} In conditional generation tasks (e.g., generating images from class labels), standard Flow Matching may struggle to distinguish between different conditions, leading to "averaged" or less distinct outputs. Contrastive Flow-Matching ($\Delta FM$)~\citep{contrastive-flowmatching} addresses this by introducing an idea from contrastive learning. It forces the model not only to learn to flow towards the correct target but also to explicitly deviate from incorrect ones.

During training, given a ``positive'' pair ($x_0, y, \epsilon$) and a ``negative'' pair ($\hat{x_0}, \hat{\epsilon}$) from the same batch, the goal of $\Delta FM$ is two-fold: (1) Push the flow at state $x_t$ closer to the positive target vector, $v^+ = \epsilon - x_0$. (2) Simultaneously, push it away from the negative target vector, $v^- = \hat{\epsilon} - \hat{x_0}$. This is achieved with a loss function that combines these two objectives:
\begin{equation}
\begin{split}
    \mathcal{L}_{\Delta FM}(\theta) = \mathbb{E}\big[&\left\|v_{\theta}(x_t, t, y) - v^+ \right\|^2 \\
    &- \lambda \left\|v_{\theta}(x_t, t, y) - v^- \right\|^2 \big]
    \label{ContrastiveFM}
\end{split}
\end{equation}
Here, $x_t = (1-t)x_0 + t\epsilon$, $\lambda$ is a hyperparameter  (set to 0.05 in our experiments) that controls the strength of the repulsive force. By encouraging greater separation between flows for different conditions, $\Delta FM$ learns a more discriminative generative process, improving sample quality and diversity.

In our model training, we used Contrastive Flow-Matching~\citep{contrastive-flowmatching} to accelerate learning in the early pre-training stage. In later stages, we found it offered no significant benefit with small batch sizes. For efficiency, we switched to using only Conditional Flow Matching. Details are in the ``Training Strategy" section below.

\begin{table*}[t]
  \centering
  \caption{Comparison on Image Generation benchmarks. ``Gen. Only'' refers to pure generation models, while ``Unified'' indicates models capable of both understanding and generation. `*' refers to the methods using LLM rewriter.}
  \vspace{-0.5em}
  \fontsize{7.5pt}{7.5pt}\selectfont
  \begin{tabular}{cll|ccc|ccc}
    \toprule
    \multirow{2}{*}{\textbf{Type}} & \multirow{2}{*}{\textbf{Method}} & \multirow{2}{*}{\textbf{Size}} & \multicolumn{3}{c|}{\textbf{Geneval}} & \multicolumn{3}{c}{\textbf{DPG-Bench}} \\
    \cmidrule(lr){4-6} \cmidrule(lr){7-9}
    ~ & ~ & ~ & \textbf{Position$\uparrow$} & \textbf{Color Attr.$\uparrow$} & \textbf{Overall$\uparrow$} & \textbf{Global$\uparrow$} & \textbf{Entity$\uparrow$} & \textbf{Overall$\uparrow$}  \\
    \midrule
    \multirow{2}{*}{\shortstack{Gen. Only}} &  FLUX.1-dev~\citep{fluxdev} & 12B & 0.20 & 0.47 & 0.67 & 82.1 & 89.5 & 84.0 \\
    ~ & SD3-Medium~\citep{sd3} & 2B & 0.33 & 0.60 & 0.74 & 87.90 & 91.01 & 84.08 \\
    \midrule
    \multirow{8}{*}{\shortstack{Unified}} & EMU3.5~\citep{emu3.5} & 34B & -- & -- & 0.86 & -- & -- & 88.26 \\
    ~ & QWen-Image~\citep{qwenimage} & 27B & 0.76 & 0.77 & 0.87 & \textbf{91.32} & \textbf{91.56} & \textbf{88.32} \\
    ~ & Bagel$^{*}$~\citep{bagel} & 7B & 0.78 & 0.77 & \textbf{0.88} & 88.94 & 90.37 & 85.07 \\
    ~ & OmniGen2$^{*}$~\citep{omnigen2} & 7B & 0.71 & 0.75 & 0.86 & 88.81 & 88.83 & 83.57 \\
    ~ & UniWorld-V1$^{*}$~\citep{uniworld} & 8B & 0.74 & 0.71 & 0.84 & 83.64 & 88.39 & 81.38 \\
    ~ & Query-Kontext$^{*}$~\citep{querykontext} & 17B & 0.85 & \textbf{0.79} & \textbf{0.88} & -- & -- & -- \\
    ~ & MetaQuery-XL$^{*}$~\citep{metaquery} & 9B & -- & -- & 0.80 & -- & -- & 82.05 \\
    ~ & Bifrost-1~\citep{bifrost} & 19B & -- & -- & 0.81 & -- & -- & 77.67 \\
    \rowcolor{gray!25}
    ~ & WeMMU (Stage 3) & 8B & \textbf{0.86} & 0.77 & \textbf{0.88} & 87.46 & 89.37 & 83.69\\
    \rowcolor{gray!25}
    ~ & WeMMU (Stage 4) & 8B & 0.85 & 0.78 & \textbf{0.88} & 87.66 & 89.07 & 83.60\\
    \bottomrule
  \end{tabular}
  \label{Text2ImageGen}
\end{table*}

\subsection{Overall Framework}
To establish a highly efficient and task-robust bridge between pretrained models, we adopt a principle of labor division: allowing each model to perform its specialized task while using query tokens as the information medium. The Vision-Language Model (VLM) understands the image and follows text instructions to consolidate essential information needed for generating the target image. The Diffusion Model then decodes this information from the VLM into pixel space. As illustrated in Fig.~\ref{fig:method}, we keep the original VLM parameters frozen to maintain its image understanding ability. At the same time, we add a new image generation pathway~\citep{bagel} with Noisy Query Tokens to integrate the information needed for generation. To add more fine-grained details, we include a VAE branch linked to the VLM through a simple linear layer. Finally, during fine-tuning with high-quality data, we unfreeze the Diffusion Model so it can adapt to the intermediate representations mapped by the Noisy Query Tokens. The following sections will explain each component in detail.

\paragraph{Noisy Query Tokens.} As the core component of our Probabilistic Expert Bridge, we introduce Noisy Query Tokens to combat the ``task generalization collapse'' where conventional learnable queries overfit to a task-specific mean representation. Instead of using fixed vectors, we inject stochasticity by sampling a new set of noisy query tokens $Q_{noisy}\sim \mathcal{N}(0, \textit{I})$ at each training step. This prevents the model from developing a path dependency on the queries and forces the VLM to learn a robust, generalizable intermediate representation distribution, which is crucial for continual learning and task scalability. For seamless integration and to support dynamic resolutions, the number of tokens is set to dynamically match the number of image patches output by the VLM's vision encoder. These tokens are then assigned specialized image-form positional embeddings (e.g., M-RoPE~\citep{qwen2vl} in Qwen2.5-VL~\citep{qwen2p5VL}) to be properly situated within the VLM's attention mechanism. We initially experimented with adding a learnable, channel-wise scaling factor to the noise. However, we found it provided no benefit: after training on nearly 80M samples, the factors remained stable around 1.0 (mean 1.0, std 0.0074), and removing them caused no performance degradation. This confirmed that direct sampling from a standard normal distribution is a simple yet robustly effective strategy.

\paragraph{VAE Branch.} To supplement the fine-grained visual details lost during the VLM's semantic processing, we introduce a VAE branch. The necessity of this component was revealed in a preliminary experiment: while raw ViT image features from Qwen2.5-VL~\citep{qwen2p5VL} could directly guide a diffusion model (Sana 1.6B) to perform high-fidelity reconstruction, routing these features through the LLM with query tokens—our core generative pathway—caused the reconstruction to collapse to a semantic-level approximation. This indicates that the VLM's processing, inherently biased towards semantic understanding, inadvertently filters out the high-frequency details present in the ViT embeddings. Consequently, to reduce the difficulty of detail preservation for the VLM, we inject features from a frozen VAE encoder (the one in the diffusion model) into the LLM via a linear layer to provide the necessary high-frequency details. To ensure seamless integration, the VAE feature length is matched to the ViT's by resizing the input image, and both are assigned identical positional embeddings. We ablate various connection methods in the Experiments section.

\paragraph{Position MLP.} To align the feature dimensions between the VLM and diffusion model and to inject explicit spatial cues, we introduce a Position MLP. The module first enhances the VLM's output features by superimposing a learnable 2D absolute positional embedding. To support dynamic resolutions, this positional embedding is dynamically cropped from the center of a large, pre-defined matrix to match the spatial dimensions of the input feature map, thereby providing a consistent spatial reference for inputs of different resolutions. Subsequently, a simple Multi-Layer Perceptron (MLP) projects these position-aware features to the target dimension required by the diffusion model's conditioner. See the Appendix~\ref{app:positionmlp} for detailed parameters.

\paragraph{VLM and Diffusion Model.} Our framework integrates a frozen pretrained VLM (Qwen2.5-VL-3B) for understanding and a tunable Diffusion Model (Sana 1.6B~\citep{sana1p5}) for generation, selected for their dynamic resolution support and performance-efficiency balance. To adapt the VLM for generation while preserving its knowledge, we employ the Expert Pathway strategy proposed in Sec.~\ref{sec:intro}. Specifically, we keep the original VLM parameters frozen and introduce a parallel, trainable generative pathway initialized from its weights. Within this pathway, Noisy Query Tokens use bidirectional attention to interact with all image and text tokens, achieving deep semantic alignment similar to native MLLMs while the original VLM tokens retain their standard attention patterns to maintain foundational behaviors. The Diffusion Model is unfrozen during fine-tuning to adapt its generative priors to the conditioning signals from the VLM's new pathway, and the overall model-agnostic design allows for future backbone upgrades.

\subsection{Training Strategy} 
Our model is trained using a four-stage progressive curriculum designed for stability, efficiency, and performance. This strategy involves gradually increasing data quality, image resolution, and task complexity, while strategically unfreezing different components and adapting the training objective.

\paragraph{Stage 1: Bridge Components Warm-up.} The primary objective of this stage is to rapidly pre-train the newly introduced bridge components. We use a large-scale, medium-quality dataset for image reconstruction and text-to-image generation tasks at a 512x512 base resolution with dynamic aspect ratios. To accelerate learning and enhance discriminative power in this high-volume data stage, we employ a Contrastive Flow Matching loss (Eq.~\ref{ContrastiveFM}). Crucially, only the VAE's linear layer, the VLM's generative pathway, and the Position MLP are trainable. Keeping the large VLM and Diffusion backbones frozen allows these smaller modules to converge to a stable state efficiently. This crucial step ensures they learn to produce a coherent and stable conditioning signal before engaging the full generative backbone, preventing the powerful diffusion model from being destabilized by chaotic initial gradients.

\begin{table*}[t]
  \centering
  \caption{Comparison on Image Editing benchmarks. ``Gen. Only'' refers to pure generation models, while ``Unified'' indicates models capable of both understanding and generation.}
  \fontsize{7.5pt}{7.5pt}\selectfont
  \begin{tabular}{cll|ccc|ccc}
    \toprule
    \multirow{2}{*}{\textbf{Type}} & \multirow{2}{*}{\textbf{Method}} & \multirow{2}{*}{\textbf{Size}} & \multicolumn{3}{c|}{\textbf{ImageEdit}} & \multicolumn{3}{c}{\textbf{GEdit-Bench-EN}} \\
    \cmidrule(lr){4-6} \cmidrule(lr){7-9}
    ~ & ~ & ~ & \textbf{Hybrid$\uparrow$} & \textbf{Action$\uparrow$} & \textbf{Overall$\uparrow$} & \textbf{G\_SC$\uparrow$} & \textbf{G\_PQ$\uparrow$} & \textbf{G\_O$\uparrow$}  \\
    \midrule
    \multirow{1}{*}{\shortstack{Gen. Only}}  & Gemini 2.5 Flash Image~\citep{gemini2.5} & -- & 3.66 & 4.59 & 4.28 & 7.41 & \textbf{7.96} & 7.10 \\
    \midrule
    \multirow{8}{*}{\shortstack{Unified}} & EMU3.5~\citep{emu3.5} & 34B & 3.69 & 4.57 & ~\textbf{4.41} & 8.11 & 7.70 & 7.59 \\
    ~ &  GPT-4o~\citep{gpt4o} & -- & \textbf{3.96} & \textbf{4.89} & 4.2 & 7.85 & 7.62 & 7.53 \\
    ~ & QWen-Image~\citep{qwenimage} & 27B & 3.82 & 4.69 & 4.27 & 8.00 & 7.86 & 7.56 \\
    ~ & Bagel~\citep{bagel} & 7B & 2.38 & 4.17 & 3.2 & 7.36 & 6.83 & 6.52 \\
    ~ & OmniGen2~\citep{omnigen2} & 7B & 2.52 & 4.68 & 3.44 & 7.16 & 6.77 & 6.41 \\
    ~ & UniWorld-V1~\citep{uniworld} & 8B & 2.96 & 2.74 & 3.26 & 4.93 & 7.43 & 4.85 \\
    ~ & Query-Kontext~\citep{querykontext} & 17B & -- & -- & -- & \textbf{8.36} & 7.37 & \textbf{7.66} \\
    \rowcolor{gray!25}
    ~ & WeMMU (Stage 3) & 8B & 2.82 & 3.15 & 3.31 & 5.86 & 6.80 & 5.75\\
    \rowcolor{gray!25}
    ~ & WeMMU (Stage 4) & 8B & 2.78 & 3.17 & 3.30 & 5.85 & 6.79 & 5.77\\
    \bottomrule
  \end{tabular}
  \label{EditMetrics}
\end{table*}

\begin{table*}[t]
  \centering
  \caption{\textbf{Detailed Training Curriculum and Hyper-parameters.} The training progresses through four stages, adjusting resolution, batch size, and data mixture. Notably, Stage 3 and Stage 4 utilize different subsets of the Uniworld dataset to target specific editing capabilities. Task abbreviations: \textbf{Rec.} (Reconstruction), \textbf{T2I} (Text-to-Image), \textbf{Uncond.} (Unconditional T2I), \textbf{S-Edit} (Single-Image Editing), \textbf{M-Edit} (Multi-Image Editing).}
  \fontsize{8pt}{9.5pt}\selectfont
  \setlength{\tabcolsep}{4pt}
  \begin{tabular}{c|ccccccc|l}
    \toprule
    \textbf{Stage} & \textbf{Res.} & \textbf{Batch} & \textbf{Steps} & \textbf{Samples} & \textbf{LR} & \textbf{Warmup} & \textbf{Dataset Source} & \textbf{Task Mixture (Ratio)} \\
    \midrule
    \textbf{1} & $512^2$ & 2336 & 34k & $\sim$80M & $1.0e^{-4}$ & 1000 & \multirow{2}{*}{\shortstack[l]{CC12M + \\ LAION-Aesthetics}} & Rec.: 50, T2I: 47, Uncond.: 3 \\
    \cmidrule{1-7} \cmidrule{9-9}
    \textbf{2} & $1024^2$ & 584 & 44k & $\sim$25.6M & $1.0e^{-5}$ & 1500 & & Rec.: 50, T2I: 47, Uncond.: 3 \\
    \midrule
    \textbf{3} & $1024^2$ & 584 & 34k & $\sim$20M & $1.0e^{-5}$ & 1500 & \shortstack[l]{HQ Mix$^\dagger$ + \\ Uniworld-V1 (\textbf{Single})} & \shortstack[l]{S-Edit: 55, T2I: 25, Rec.: 20, \\ Uncond.: 10} \\
    \cmidrule{1-9}
    \textbf{4} & $1024^2$ & 244 & 18k & $\sim$4.3M & $1.0e^{-5}$ & 1500 & \shortstack[l]{HQ Mix$^\dagger$ + \\ Uniworld-V1 (\textbf{Multi})} & \shortstack[l]{S-Edit: 35, M-Edit: 20, T2I: 25, \\ Rec.: 20, Uncond.: 10} \\
    \bottomrule
    \multicolumn{9}{l}{\scriptsize $^\dagger$ HQ Mix includes Blip3o, shareGPT-4o, and OpenGPT-4o.}
  \end{tabular}
  \label{tab:training_details}
\end{table*}

\paragraph{Stage 2: Joint Adaptation with Diffusion Model.} With the bridge now providing a stable signal, we proceed to adapt the diffusion model. Using the same dataset, we increase the base resolution to 1024x1024 and unfreeze the entire Sana diffusion model. The higher resolution necessitates smaller batch sizes. Since the effectiveness of Contrastive Flow Matching relies on a rich diversity of in-batch negative samples, its benefits diminish significantly in a small-batch regime. Therefore, we switch to the standard Conditional Flow Matching loss (Eq.~\ref{ConditionFM}) for all subsequent stages. This stage establishes the foundational alignment between the VLM's conditioning signals and Sana's generative priors at a high resolution.

\paragraph{Stage 3: High-Fidelity Generation and Core Task Learning.} Leveraging the high-resolution alignment achieved in the previous stage, we now shift the focus to refining generation quality and mastering our core task of single-image editing. To this end, we introduce a high-quality curated dataset and adjust the task mixture to prioritize high-fidelity text-to-image generation and editing instructions. A small amount of reconstruction data is retained to maintain fidelity, and all previously trainable parameters continue to be optimized.

\paragraph{Stage 4: Generalization to Complex Tasks.} The final stage aims to validate and enhance the model's task generalization. While maintaining the text-to-image task, we reduce the proportion of single-image editing data and introduce more complex multi-image editing tasks. This demonstrates that our framework can be extended to new, challenging tasks without catastrophic forgetting, validating the effectiveness of our Noisy Query Token approach.

\section{Experiment}
\label{sec:experiment}
\subsection{Dataset and Implementation Details}
Our model is built upon the pre-trained Qwen2.5-VL-3B~\citep{qwen2p5VL} and Sana 1.6B~\citep{sana1p5} architectures. The training process employs the AdamW optimizer with $\beta_1=0.9$ and $\beta_2=0.95$. To achieve robust instruction following and high-fidelity generation, we designed a four-stage curriculum learning strategy. This curriculum progressively scales up image resolution and task complexity, evolving from basic reconstruction to complex multi-image editing.

\begin{figure*}
    \centering
    \includegraphics[width=0.95\linewidth]{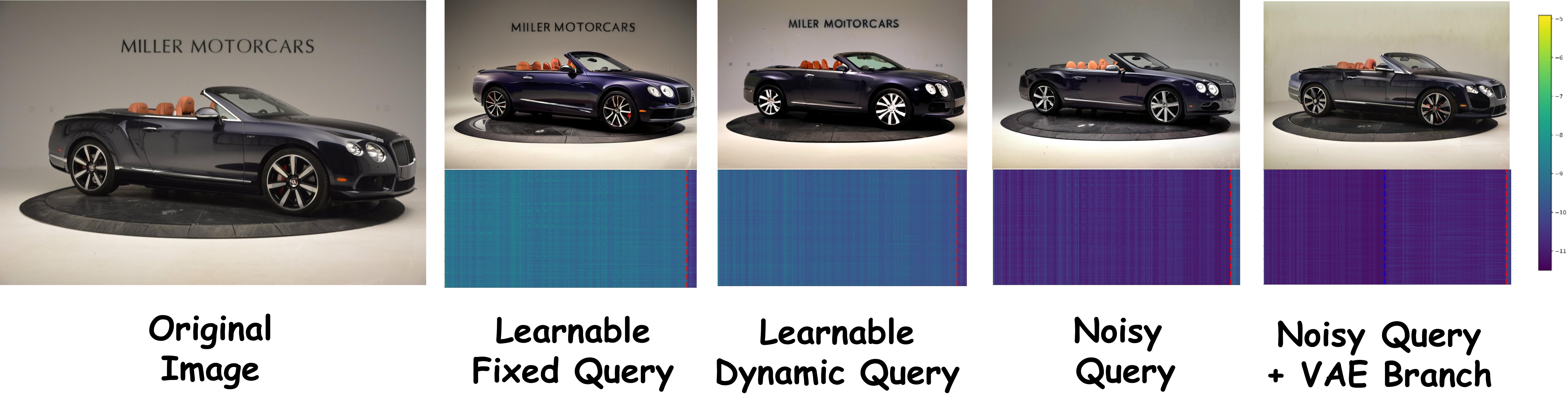}
    \caption{Analysis of query token attention mechanisms. The prompt is ``Remove the `MILLER MOTORCARS' text positioned across the top center of the image''. (Bottom rows) Learnable queries show a strong attention bias towards image tokens. Our Noisy Queries shift focus to the text tokens (right of red line), prioritizing instruction following. The VAE branch (right of blue line) helps balance this attention.}
    \label{fig:noisyaba}
\end{figure*}

Detailed configurations for each training stage are summarized in Table~\ref{tab:training_details}.
In the initial stages (Stage 1-2), we utilize large-scale public datasets, CC12M~\citep{cc12m} and LAION-aesthetics-12M, with captions recaptioned by InternVL2-8B~\citep{internVL_8b}. For the advanced stages (Stage 3-4), we curate a high-quality dataset mix (HQ Mix) from Blip3o~\citep{blip3o}, shareGPT-4o~\citep{sharegpt4o}, and OpenGPT-4o~\citep{opengpt4o}. Regarding editing data, Stage 3 utilizes the single-image editing subset from Uniworld-V1~\citep{uniworld}, while Stage 4 shifts focus to the multi-reference image editing subset to enable complex composition capabilities.

\subsection{Comparison with Other Systems}
We evaluated the performance of our model after the completion of Stage 3 and Stage 4 training using the GenEval~\citep{GenEval} and DPG-Bench~\citep{DPGBench} benchmarks. As shown in Table~\ref{Text2ImageGen}, our model achieves the highest score on GenEval among models not fine-tuned with Reinforcement Learning (RL), while also demonstrating strong performance on DPG-Bench.

We further assessed its editing capabilities on ImageEdit-Bench~\citep{imageedit} and GEdit-EN~\citep{gedit}, with the metrics reported in Table~\ref{EditMetrics}. Despite limitations in the quantity and quality of our editing data, our model achieves performance comparable to that of Bagel and Uniworld-V1 on ImageEdit-Bench. On GEdit-EN, our model surpasses Uniworld-V1 on the G\_SC metric and G\_O metric, while trailing on the G\_PQ metric. We present a comprehensive collection of samples generated and edited by our model in the Appendix ~\ref{app:moreimages}.

\begin{figure}
    \centering
    \includegraphics[width=0.90\linewidth]{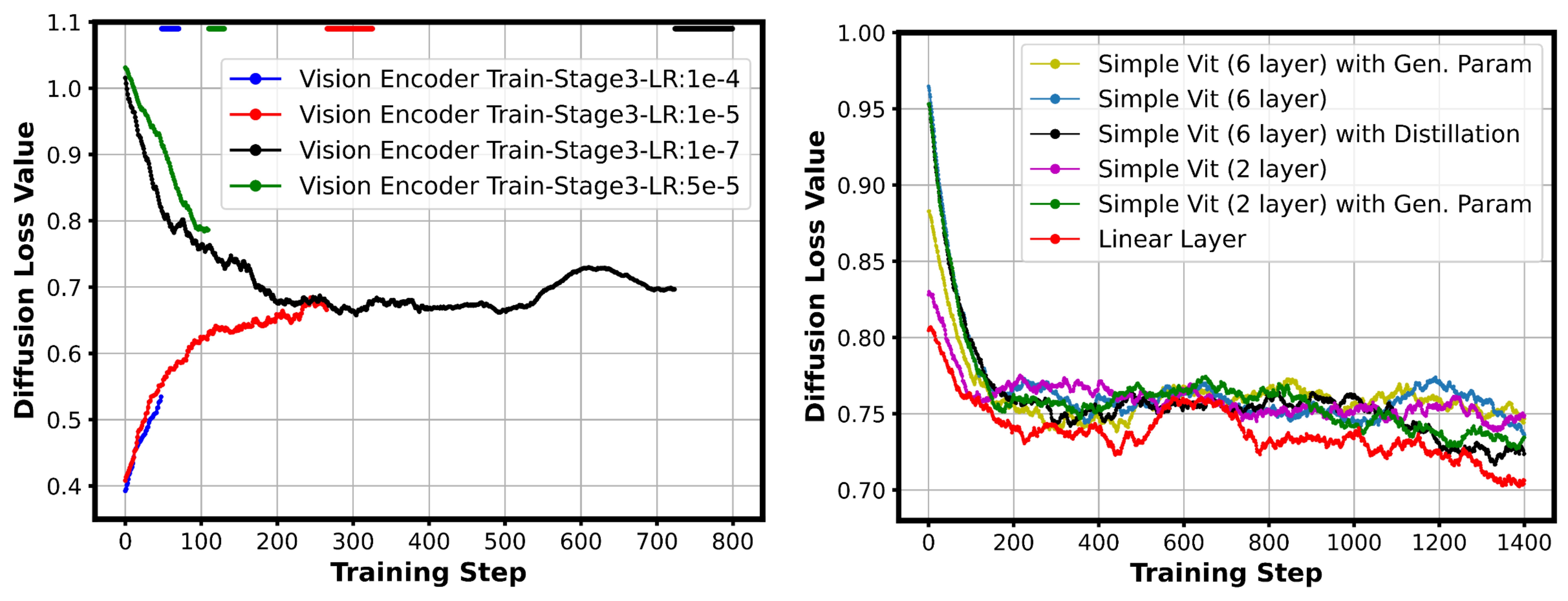}
    \caption{Justifying the VAE Branch design. (Left) Fine-tuning the native ViT of Qwen2.5-VL leads to training collapse. (Right) Loss curves for different VAE branch connection methods, showing a simple Linear layer (the red line) provides the fastest and most stable convergence.}
    \label{fig:vaeloss}
    \vspace{-1.5em}
\end{figure}

\subsection{Ablation Study}

\begin{figure*}
    \centering
    \vspace{-1em}
    \includegraphics[width=0.95\linewidth]{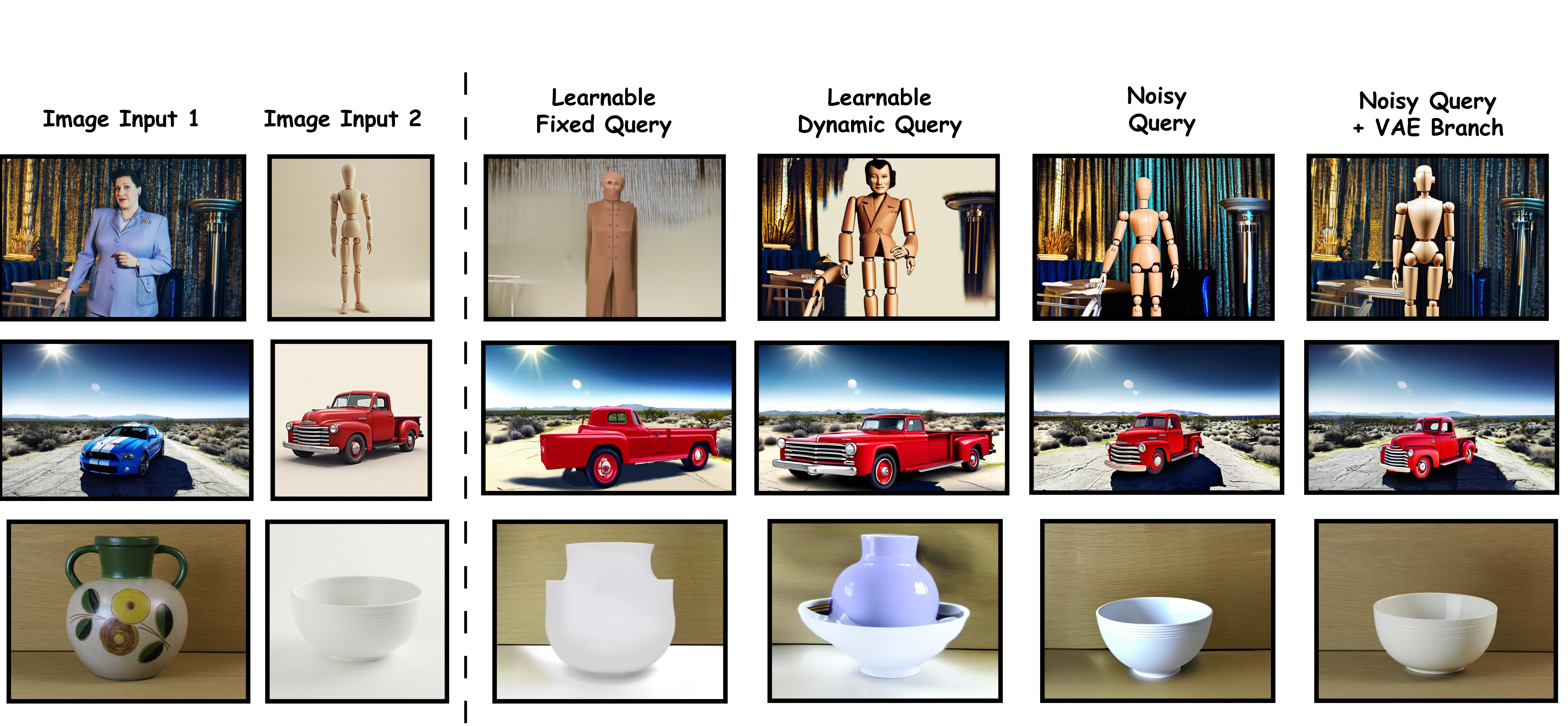}
    \caption{Generalization on the multi-image editing task. The task is to replace the subject in Input 1 with the subject from Input 2. Baselines with learnable queries (third and fourth columns) produce incoherent results. Our Noisy Query method (fifth column) correctly performs the edit, while our full model (last column) improves detail fidelity.}
    \label{fig:multiimg}
\end{figure*}

\subsubsection{Ablation on Query Token Design}
To comprehensively evaluate the task generalization capabilities of our Noisy Query Token against conventional learnable queries, we designed a rigorous ablation study. We established four experimental settings: (1) \textbf{Learnable Fixed Query:} A standard baseline with a fixed number of learnable query tokens. (2) \textbf{Learnable Dynamic Query:} An enhanced baseline where the learnable query tokens are dynamically sampled based on the input resolution, improving their adaptability to varying spatial dimensions. (3) \textbf{Noisy Query:} Our proposed method, using query tokens sampled from a standard normal distribution at each step, without the VAE branch. (4) \textbf{Noisy Query + VAE Branch:} Our full proposed model. All four configurations were trained under identical conditions up to the end of Stage 3.

\textbf{Quantitative Results.} The performance on the ImageEdit-Bench, presented in Table~\ref{Tab:QueryAblation}, clearly demonstrates the superiority of our proposed components. While making the learnable queries dynamic provides a modest improvement over the fixed baseline, a significant leap in performance is observed when we replace them entirely with our Noisy Query tokens. The inclusion of the VAE Branch further boosts all metrics, leading to the best overall performance and confirming the efficacy of each component in our final design.

\textbf{Analysis of Attention Mechanism.} To understand the underlying mechanism behind these performance gains, we visualized the attention maps from the query tokens to all image and text tokens, as shown in Fig.~\ref{fig:noisyaba}. (In the attention maps, a red line separates image tokens from text tokens, and a blue line separates the original ViT features from the injected VAE features). A quantitative analysis reveals a critical insight: we calculated the difference between the average attention paid to image tokens versus text tokens for each setup, yielding values of 1.80, 1.01, -0.99, and -0.68, respectively.

This reveals a fundamental difference in behavior: both Learnable Query variants show a strong attention bias towards image tokens. In stark contrast, our Noisy Query method shifts its focus towards the text tokens. This suggests that Noisy Tokens prioritize understanding and following the textual instruction for editing, rather than simply reconstructing the source image. Interestingly, the addition of the VAE Branch, while further improving scores, also brings the attention gap between the two modalities closer (from -0.99 to -0.68), suggesting it helps create a more balanced representation by offloading some of the fine-grained visual reconstruction burden.

\begin{table}[t]
  \centering
  \setlength{\tabcolsep}{4pt} 
  \caption{Ablation Study on Query Token Design using the ImageEdit Benchmark}
  \vspace{-0.5em}
  \fontsize{7.0pt}{7.0pt}\selectfont
  \begin{tabular}{l|ccc}
    \toprule
     \textbf{Exp. Setting} &
     \textbf{Hybrid}$\uparrow$ & \textbf{Action}$\uparrow$ & \textbf{Overall}$\uparrow$ \\
    \midrule
     ~ Learnable Fixed Query & 1.87  & 2.21  & 2.53\\
    \midrule
    ~ Learnable Dynamic Query & 2.02  & 2.60 & 2.88 \\
    \midrule
    ~ Noisy Query & 2.36 & 2.75 & 2.98\\
    \midrule
    ~ Noisy Query + VAE Branch & \textbf{2.82}  & \textbf{3.15}  & \textbf{3.31}\\
    \bottomrule
  \end{tabular}
  \label{Tab:QueryAblation}
\end{table}

\subsubsection{Ablating VAE Branch Integration}
A key challenge in our framework is preserving the fine-grained visual details that can be lost during the VLM's high-level semantic processing. To address this, we first explored the most direct solution: unfreezing the VLM's own Vision Encoder (ViT) to allow it to adapt. While this approach led to near-perfect image reconstruction, we discovered it was highly unstable during editing-task fine-tuning. As shown in Fig.~\ref{fig:vaeloss} (left), fine-tuning the ViT of Qwen2.5-VL consistently resulted in catastrophic training collapse, regardless of hyperparameter adjustments. Interestingly, we observed that this instability is model-dependent; a smaller LLaVA-OV 0.5B~\citep{llava_ov} model with a SigLIP ViT~\citep{siglip} could be successfully fine-tuned without collapse. We hypothesize that the pre-training objectives and architecture of the ViT heavily influence its adaptability. This led us to conclude that unfreezing the native Vision Encoder is not a robust or generalizable strategy for injecting detail.

To avoid the instability of modifying core VLM components, we introduced a dedicated branch with a frozen, powerful VAE encoder. The challenge was integrating its detail-rich features into the LLM. We explored six connection methods: (1) A simple Linear Layer, (2) A lightweight, 2-layer Vision Transformer (ViT) (118M parameters). (3) A 2-layer ViT, then processed by the VLM's generative pathway. (4) A 6-layer ViT (319M parameters) (5) A 6-layer ViT then processed by the VLM's generative pathway. (6) A 6-layer ViT trained with distillation from the Qwen2.5-VL's ViT.
As illustrated by the training loss curves in Fig.~\ref{fig:vaeloss} (right), the simplest approach—a single Linear Layer—demonstrated the fastest and most stable convergence. This result suggests that a direct, low-parameter channel for VAE features is the most effective method, providing the necessary high-frequency details without introducing unnecessary complexity that could interfere with the primary learning objective. This confirms our final architectural choice.

\subsection{Generalization to Multi-Image Tasks}
Stage 4 was designed to test the adaptability of our \textbf{Noisy Query Token} approach on a complex new task: multi-image editing. This task serves as a powerful litmus test for generalization, as it requires the model to fuse information from multiple sources in a way it has not been explicitly trained for until this stage. To test for generalization collapse, all four model configurations (trained through Stage 3) were subsequently fine-tuned on the Stage 4 curriculum.

The qualitative results, vividly illustrated in Fig.~\ref{fig:multiimg}, reveal a stark difference in the generalization capabilities of the query mechanisms. The baselines using both Learnable Fixed and Learnable Dynamic queries fundamentally fail to grasp the task, producing chaotic and incoherently fused images. In stark contrast, the model equipped with Noisy Query tokens successfully follows the instructions to perform the correct multi-image replacement. Our full model, which incorporates the VAE Branch, further enhances the textural details and fidelity of the edited region. While these results prove the scalability of our framework, we note that even our best outputs can exhibit minor visual artifacts at composite seams, suggesting that higher-quality data or RL-based fine-tuning are promising directions for future refinement.

Most importantly, this stage validates the core advantage of our method: the model can stably learn new, complex tasks without suffering from catastrophic forgetting. To verify this, we re-evaluated the model on its prior tasks and found that its performance in \textbf{both single-image editing and text-to-image generation} was preserved after Stage 4 training, refer to Table~\ref{Text2ImageGen} and Table~\ref{EditMetrics}. The ability to acquire new skills while retaining existing ones provides strong evidence for the robustness and superior generalization capabilities endowed by our approach.

\section{Discussion and Conclusion}
\label{sec:conclusion}
We identify task generalization collapse as a critical flaw in learnable query tokens for bridging VLMs and diffusion models. Our solution, Noisy Query Tokens, samples queries from a standard normal distribution at each step, forcing the model to learn a robust intermediate representation space rather than task-specific shortcuts. Complemented by a VAE branch that preserves fine-grained details via linear projection, our WeMMU framework achieves state-of-the-art editing performance and competitive generation results while enabling stable continual learning. Minor artifacts in multi-image editing suggest room for data quality improvements or RL fine-tuning. Future directions include scaling to larger backbones and exploring adaptive noise schedules. WeMMU offers a simple, effective paradigm for sustainable multimodal unification.
{
    \small
    \bibliographystyle{ieeenat_fullname}
    \bibliography{main}

\begin{thebibliography}{44}
\providecommand{\natexlab}[1]{#1}
\providecommand{\url}[1]{\texttt{#1}}
\expandafter\ifx\csname urlstyle\endcsname\relax
  \providecommand{\doi}[1]{doi: #1}\else
  \providecommand{\doi}{doi: \begingroup \urlstyle{rm}\Url}\fi

\bibitem[An et~al.(2025)An, Xie, Yang, Zhang, Zhao, Cheng, Wang, Xu, Chen, Wu, et~al.]{llava_ov}
Xiang An, Yin Xie, Kaicheng Yang, Wenkang Zhang, Xiuwei Zhao, Zheng Cheng, Yirui Wang, Songcen Xu, Changrui Chen, Chunsheng Wu, et~al.
\newblock Llava-onevision-1.5: Fully open framework for democratized multimodal training.
\newblock \emph{arXiv preprint arXiv:2509.23661}, 2025.

\bibitem[Bai et~al.(2025)Bai, Chen, Liu, Wang, Ge, Song, Dang, Wang, Wang, Tang, et~al.]{qwen2p5VL}
Shuai Bai, Keqin Chen, Xuejing Liu, Jialin Wang, Wenbin Ge, Sibo Song, Kai Dang, Peng Wang, Shijie Wang, Jun Tang, et~al.
\newblock Qwen2. 5-vl technical report.
\newblock \emph{arXiv preprint arXiv:2502.13923}, 2025.

\bibitem[Changpinyo et~al.(2021)Changpinyo, Sharma, Ding, and Soricut]{cc12m}
Soravit Changpinyo, Piyush Sharma, Nan Ding, and Radu Soricut.
\newblock Conceptual 12m: Pushing web-scale image-text pre-training to recognize long-tail visual concepts.
\newblock In \emph{{CVPR}}, pages 3558--3568. Computer Vision Foundation / {IEEE}, 2021.

\bibitem[Chen et~al.(2025{\natexlab{a}})Chen, Cai, Chen, Chen, Ji, Wang, Yang, and Wang]{sharegpt4o}
Junying Chen, Zhenyang Cai, Pengcheng Chen, Shunian Chen, Ke Ji, Xidong Wang, Yunjin Yang, and Benyou Wang.
\newblock Sharegpt-4o-image: Aligning multimodal models with gpt-4o-level image generation.
\newblock \emph{arXiv preprint arXiv:2506.18095}, 2025{\natexlab{a}}.

\bibitem[Chen et~al.(2025{\natexlab{b}})Chen, Xu, Pan, Hu, Qin, Goldstein, Huang, Zhou, Xie, Savarese, et~al.]{blip3o}
Jiuhai Chen, Zhiyang Xu, Xichen Pan, Yushi Hu, Can Qin, Tom Goldstein, Lifu Huang, Tianyi Zhou, Saining Xie, Silvio Savarese, et~al.
\newblock Blip3-o: A family of fully open unified multimodal models-architecture, training and dataset.
\newblock \emph{arXiv preprint arXiv:2505.09568}, 2025{\natexlab{b}}.

\bibitem[Chen et~al.(2025{\natexlab{c}})Chen, Wu, Liu, Pan, Liu, Xie, Yu, and Ruan]{janus-pro}
Xiaokang Chen, Zhiyu Wu, Xingchao Liu, Zizheng Pan, Wen Liu, Zhenda Xie, Xingkai Yu, and Chong Ruan.
\newblock Janus-pro: Unified multimodal understanding and generation with data and model scaling.
\newblock \emph{CoRR}, 2025{\natexlab{c}}.

\bibitem[Chen et~al.(2024)Chen, Wu, Wang, Su, Chen, Xing, Zhong, Zhang, Zhu, Lu, Li, Luo, Lu, Qiao, and Dai]{internVL_8b}
Zhe Chen, Jiannan Wu, Wenhai Wang, Weijie Su, Guo Chen, Sen Xing, Muyan Zhong, Qinglong Zhang, Xizhou Zhu, Lewei Lu, Bin Li, Ping Luo, Tong Lu, Yu Qiao, and Jifeng Dai.
\newblock Internvl: Scaling up vision foundation models and aligning for generic visual-linguistic tasks, 2024.

\bibitem[Chen et~al.(2025{\natexlab{d}})Chen, Bai, Shi, Fu, Zhang, Wang, Sun, Zhang, Wang, Zhang, et~al.]{opengpt4o}
Zhihong Chen, Xuehai Bai, Yang Shi, Chaoyou Fu, Huanyu Zhang, Haotian Wang, Xiaoyan Sun, Zhang Zhang, Liang Wang, Yuanxing Zhang, et~al.
\newblock Opengpt-4o-image: A comprehensive dataset for advanced image generation and editing.
\newblock \emph{arXiv preprint arXiv:2509.24900}, 2025{\natexlab{d}}.

\bibitem[Cui et~al.(2025)Cui, Chen, Deng, Huang, Li, Liu, Liu, Luo, Wang, Wang, et~al.]{emu3.5}
Yufeng Cui, Honghao Chen, Haoge Deng, Xu Huang, Xinghang Li, Jirong Liu, Yang Liu, Zhuoyan Luo, Jinsheng Wang, Wenxuan Wang, et~al.
\newblock Emu3. 5: Native multimodal models are world learners.
\newblock \emph{arXiv preprint arXiv:2510.26583}, 2025.

\bibitem[Deng et~al.(2025)Deng, Zhu, Li, Gou, Li, Wang, Zhong, Yu, Nie, Song, et~al.]{bagel}
Chaorui Deng, Deyao Zhu, Kunchang Li, Chenhui Gou, Feng Li, Zeyu Wang, Shu Zhong, Weihao Yu, Xiaonan Nie, Ziang Song, et~al.
\newblock Emerging properties in unified multimodal pretraining.
\newblock \emph{arXiv preprint arXiv:2505.14683}, 2025.

\bibitem[Esser et~al.(2024)Esser, Kulal, Blattmann, Entezari, M{\"u}ller, Saini, Levi, Lorenz, Sauer, Boesel, et~al.]{sd3}
Patrick Esser, Sumith Kulal, Andreas Blattmann, Rahim Entezari, Jonas M{\"u}ller, Harry Saini, Yam Levi, Dominik Lorenz, Axel Sauer, Frederic Boesel, et~al.
\newblock Scaling rectified flow transformers for high-resolution image synthesis.
\newblock In \emph{Forty-first international conference on machine learning}, 2024.

\bibitem[Ghosh et~al.(2023)Ghosh, Hajishirzi, and Schmidt]{GenEval}
Dhruba Ghosh, Hanna Hajishirzi, and Ludwig Schmidt.
\newblock Geneval: An object-focused framework for evaluating text-to-image alignment, 2023.

\bibitem[Hu et~al.(2024)Hu, Wang, Fang, Fu, Cheng, and Yu]{DPGBench}
Xiwei Hu, Rui Wang, Yixiao Fang, Bin Fu, Pei Cheng, and Gang Yu.
\newblock Ella: Equip diffusion models with llm for enhanced semantic alignment.
\newblock \emph{arXiv preprint arXiv:2403.05135}, 2024.

\bibitem[Li et~al.(2025)Li, Peng, Wang, Peng, Chen, Weng, Wang, Cai, Dai, and Xiong]{onecat}
Han Li, Xinyu Peng, Yaoming Wang, Zelin Peng, Xin Chen, Rongxiang Weng, Jingang Wang, Xunliang Cai, Wenrui Dai, and Hongkai Xiong.
\newblock Onecat: Decoder-only auto-regressive model for unified understanding and generation.
\newblock \emph{arXiv preprint arXiv:2509.03498}, 2025.

\bibitem[Liao et~al.(2025)Liao, Liu, Wang, Luo, Zhang, Zhao, Wu, Li, Tian, and Huang]{mogao}
Chao Liao, Liyang Liu, Xun Wang, Zhengxiong Luo, Xinyu Zhang, Wenliang Zhao, Jie Wu, Liang Li, Zhi Tian, and Weilin Huang.
\newblock Mogao: An omni foundation model for interleaved multi-modal generation.
\newblock \emph{arXiv preprint arXiv:2505.05472}, 2025.

\bibitem[Lin et~al.(2025{\natexlab{a}})Lin, Li, Cheng, Niu, Ye, He, Yuan, Yu, Wang, Ge, et~al.]{uniworld}
Bin Lin, Zongjian Li, Xinhua Cheng, Yuwei Niu, Yang Ye, Xianyi He, Shenghai Yuan, Wangbo Yu, Shaodong Wang, Yunyang Ge, et~al.
\newblock Uniworld: High-resolution semantic encoders for unified visual understanding and generation.
\newblock \emph{arXiv preprint arXiv:2506.03147}, 2025{\natexlab{a}}.

\bibitem[Lin et~al.(2025{\natexlab{b}})Lin, Cho, Zadeh, Li, and Bansal]{bifrost}
Han Lin, Jaemin Cho, Amir Zadeh, Chuan Li, and Mohit Bansal.
\newblock Bifrost-1: Bridging multimodal llms and diffusion models with patch-level clip latents.
\newblock \emph{arXiv preprint arXiv:2508.05954}, 2025{\natexlab{b}}.

\bibitem[Lipman et~al.()Lipman, Chen, Ben-Hamu, Nickel, and Le]{flowmatching}
Yaron Lipman, Ricky~TQ Chen, Heli Ben-Hamu, Maximilian Nickel, and Matthew Le.
\newblock Flow matching for generative modeling.
\newblock In \emph{The Eleventh International Conference on Learning Representations}.

\bibitem[Liu et~al.(2025)Liu, Han, Xing, Yin, Wang, Cheng, Liao, Wang, Fu, Han, et~al.]{gedit}
Shiyu Liu, Yucheng Han, Peng Xing, Fukun Yin, Rui Wang, Wei Cheng, Jiaqi Liao, Yingming Wang, Honghao Fu, Chunrui Han, et~al.
\newblock Step1x-edit: A practical framework for general image editing.
\newblock \emph{arXiv preprint arXiv:2504.17761}, 2025.

\bibitem[Ma et~al.(2025)Ma, Liu, Chen, Liu, Wu, Wu, Pan, Xie, Zhang, Yu, et~al.]{janusflow}
Yiyang Ma, Xingchao Liu, Xiaokang Chen, Wen Liu, Chengyue Wu, Zhiyu Wu, Zizheng Pan, Zhenda Xie, Haowei Zhang, Xingkai Yu, et~al.
\newblock Janusflow: Harmonizing autoregression and rectified flow for unified multimodal understanding and generation.
\newblock In \emph{Proceedings of the Computer Vision and Pattern Recognition Conference}, pages 7739--7751, 2025.

\bibitem[OpenAI(2025)]{gpt4o}
OpenAI.
\newblock Gpt-4o, 2025.

\bibitem[Pan et~al.(2025)Pan, Shukla, Singh, Zhao, Mishra, Wang, Xu, Chen, Li, Juefei-Xu, et~al.]{metaquery}
Xichen Pan, Satya~Narayan Shukla, Aashu Singh, Zhuokai Zhao, Shlok~Kumar Mishra, Jialiang Wang, Zhiyang Xu, Jiuhai Chen, Kunpeng Li, Felix Juefei-Xu, et~al.
\newblock Transfer between modalities with metaqueries.
\newblock \emph{arXiv preprint arXiv:2504.06256}, 2025.

\bibitem[Ronneberger et~al.(2015)Ronneberger, Fischer, and Brox]{u-net}
Olaf Ronneberger, Philipp Fischer, and Thomas Brox.
\newblock U-net: Convolutional networks for biomedical image segmentation.
\newblock In \emph{International Conference on Medical image computing and computer-assisted intervention}, pages 234--241. Springer, 2015.

\bibitem[Shi et~al.(2024)Shi, Han, Zhou, Liang, Lin, Zettlemoyer, and Yu]{lmfusion}
Weijia Shi, Xiaochuang Han, Chunting Zhou, Weixin Liang, Xi~Victoria Lin, Luke Zettlemoyer, and Lili Yu.
\newblock Lmfusion: Adapting pretrained language models for multimodal generation.
\newblock \emph{arXiv preprint arXiv:2412.15188}, 2024.

\bibitem[Song et~al.(2025)Song, Dong, Wang, Zhang, Xue, Yuan, Yang, Feng, Zhou, Xiao, et~al.]{querykontext}
Yuxin Song, Wenkai Dong, Shizun Wang, Qi Zhang, Song Xue, Tao Yuan, Hu Yang, Haocheng Feng, Hang Zhou, Xinyan Xiao, et~al.
\newblock Query-kontext: An unified multimodal model for image generation and editing.
\newblock \emph{arXiv preprint arXiv:2509.26641}, 2025.

\bibitem[Stoica et~al.(2025)Stoica, Ramanujan, Fan, Farhadi, Krishna, and Hoffman]{contrastive-flowmatching}
George Stoica, Vivek Ramanujan, Xiang Fan, Ali Farhadi, Ranjay Krishna, and Judy Hoffman.
\newblock Contrastive flow matching.
\newblock \emph{arXiv preprint arXiv:2506.05350}, 2025.

\bibitem[Team(2024)]{chameleon}
Chameleon Team.
\newblock Chameleon: Mixed-modal early-fusion foundation models.
\newblock \emph{arXiv preprint arXiv:2405.09818}, 2024.

\bibitem[Team(2025)]{gemini2.5}
Gemini Team.
\newblock Gemini 2.5 flash \& gemini 2.5 flash image model card.
\newblock 2025.

\bibitem[Wang et~al.(2025)Wang, Zhao, Zhang, Cao, Zhan, Duan, Lu, Fu, Chen, Zhao, et~al.]{ovis-u1}
Guo-Hua Wang, Shanshan Zhao, Xinjie Zhang, Liangfu Cao, Pengxin Zhan, Lunhao Duan, Shiyin Lu, Minghao Fu, Xiaohao Chen, Jianshan Zhao, et~al.
\newblock Ovis-u1 technical report.
\newblock \emph{arXiv preprint arXiv:2506.23044}, 2025.

\bibitem[Wang et~al.(2024{\natexlab{a}})Wang, Bai, Tan, Wang, Fan, Bai, Chen, Liu, Wang, Ge, et~al.]{qwen2vl}
Peng Wang, Shuai Bai, Sinan Tan, Shijie Wang, Zhihao Fan, Jinze Bai, Keqin Chen, Xuejing Liu, Jialin Wang, Wenbin Ge, et~al.
\newblock Qwen2-vl: Enhancing vision-language model's perception of the world at any resolution.
\newblock \emph{arXiv preprint arXiv:2409.12191}, 2024{\natexlab{a}}.

\bibitem[Wang et~al.(2024{\natexlab{b}})Wang, Zhang, Luo, Sun, Cui, Wang, Zhang, Wang, Li, Yu, et~al.]{emu3}
Xinlong Wang, Xiaosong Zhang, Zhengxiong Luo, Quan Sun, Yufeng Cui, Jinsheng Wang, Fan Zhang, Yueze Wang, Zhen Li, Qiying Yu, et~al.
\newblock Emu3: Next-token prediction is all you need.
\newblock \emph{CoRR}, 2024{\natexlab{b}}.

\bibitem[Wu et~al.(2025{\natexlab{a}})Wu, Chen, Wu, Ma, Liu, Pan, Liu, Xie, Yu, Ruan, et~al.]{janus}
Chengyue Wu, Xiaokang Chen, Zhiyu Wu, Yiyang Ma, Xingchao Liu, Zizheng Pan, Wen Liu, Zhenda Xie, Xingkai Yu, Chong Ruan, et~al.
\newblock Janus: Decoupling visual encoding for unified multimodal understanding and generation.
\newblock In \emph{Proceedings of the Computer Vision and Pattern Recognition Conference}, pages 12966--12977, 2025{\natexlab{a}}.

\bibitem[Wu et~al.(2025{\natexlab{b}})Wu, Li, Zhou, Lin, Gao, Yan, Yin, Bai, Xu, Chen, et~al.]{qwenimage}
Chenfei Wu, Jiahao Li, Jingren Zhou, Junyang Lin, Kaiyuan Gao, Kun Yan, Sheng-ming Yin, Shuai Bai, Xiao Xu, Yilei Chen, et~al.
\newblock Qwen-image technical report.
\newblock \emph{arXiv preprint arXiv:2508.02324}, 2025{\natexlab{b}}.

\bibitem[Wu et~al.(2025{\natexlab{c}})Wu, Zheng, Yan, Xiao, Luo, Wang, Li, Jiang, Liu, Zhou, et~al.]{omnigen2}
Chenyuan Wu, Pengfei Zheng, Ruiran Yan, Shitao Xiao, Xin Luo, Yueze Wang, Wanli Li, Xiyan Jiang, Yexin Liu, Junjie Zhou, et~al.
\newblock Omnigen2: Exploration to advanced multimodal generation.
\newblock \emph{arXiv preprint arXiv:2506.18871}, 2025{\natexlab{c}}.

\bibitem[Xie et~al.()Xie, Chen, Zhao, YU, Zhu, Lin, Zhang, Li, Chen, Cai, et~al.]{sana1p5}
Enze Xie, Junsong Chen, Yuyang Zhao, Jincheng YU, Ligeng Zhu, Yujun Lin, Zhekai Zhang, Muyang Li, Junyu Chen, Han Cai, et~al.
\newblock Sana 1.5: Efficient scaling of training-time and inference-time compute in linear diffusion transformer.
\newblock In \emph{Forty-second International Conference on Machine Learning}.

\bibitem[Xie et~al.(2025)Xie, Darrell, Zettlemoyer, and Wang]{recA}
Ji Xie, Trevor Darrell, Luke Zettlemoyer, and XuDong Wang.
\newblock Reconstruction alignment improves unified multimodal models.
\newblock \emph{arXiv preprint arXiv:2509.07295}, 2025.

\bibitem[Xu et~al.(2025)Xu, Yin, and Chen]{tbac}
Junzhe Xu, Yuyang Yin, and Xi Chen.
\newblock Tbac-uniimage: Unified understanding and generation by ladder-side diffusion tuning.
\newblock \emph{arXiv preprint arXiv:2508.08098}, 2025.

\bibitem[Yang et~al.(2024{\natexlab{a}})Yang, Liu, Deng, Kim, Mei, Shen, and Chen]{flux-dev}
Chenglin Yang, Celong Liu, Xueqing Deng, Dongwon Kim, Xing Mei, Xiaohui Shen, and Liang-Chieh Chen.
\newblock 1.58-bit flux.
\newblock \emph{arXiv preprint arXiv:2412.18653}, 2024{\natexlab{a}}.

\bibitem[Yang et~al.(2024{\natexlab{b}})Yang, Liu, Deng, Kim, Mei, Shen, and Chen]{fluxdev}
Chenglin Yang, Celong Liu, Xueqing Deng, Dongwon Kim, Xing Mei, Xiaohui Shen, and Liang-Chieh Chen.
\newblock 1.58-bit flux.
\newblock \emph{arXiv preprint arXiv:2412.18653}, 2024{\natexlab{b}}.

\bibitem[Yang et~al.(2025)Yang, Yin, Zhou, Rao, Zhai, Cao, and Zha]{mmar}
Jian Yang, Dacheng Yin, Yizhou Zhou, Fengyun Rao, Wei Zhai, Yang Cao, and Zheng-Jun Zha.
\newblock Mmar: Towards lossless multi-modal auto-regressive probabilistic modeling.
\newblock In \emph{Proceedings of the Computer Vision and Pattern Recognition Conference}, pages 7974--7985, 2025.

\bibitem[Ye et~al.(2025)Ye, He, Li, Lin, Yuan, Yan, Hou, and Yuan]{imageedit}
Yang Ye, Xianyi He, Zongjian Li, Bin Lin, Shenghai Yuan, Zhiyuan Yan, Bohan Hou, and Li Yuan.
\newblock Imgedit: A unified image editing dataset and benchmark.
\newblock \emph{arXiv preprint arXiv:2505.20275}, 2025.

\bibitem[Zhai et~al.(2023)Zhai, Mustafa, Kolesnikov, and Beyer]{siglip}
Xiaohua Zhai, Basil Mustafa, Alexander Kolesnikov, and Lucas Beyer.
\newblock Sigmoid loss for language image pre-training.
\newblock In \emph{Proceedings of the IEEE/CVF international conference on computer vision}, pages 11975--11986, 2023.

\bibitem[Zhou et~al.()Zhou, YU, Babu, Tirumala, Yasunaga, Shamis, Kahn, Ma, Zettlemoyer, and Levy]{transfusion}
Chunting Zhou, LILI YU, Arun Babu, Kushal Tirumala, Michihiro Yasunaga, Leonid Shamis, Jacob Kahn, Xuezhe Ma, Luke Zettlemoyer, and Omer Levy.
\newblock Transfusion: Predict the next token and diffuse images with one multi-modal model.
\newblock In \emph{The Thirteenth International Conference on Learning Representations}.

\bibitem[Zhuang et~al.(2025)Zhuang, Xie, Deng, Liang, Ru, Yin, and Zou]{vargpt}
Xianwei Zhuang, Yuxin Xie, Yufan Deng, Liming Liang, Jinghan Ru, Yuguo Yin, and Yuexian Zou.
\newblock Vargpt: Unified understanding and generation in a visual autoregressive multimodal large language model.
\newblock \emph{CoRR}, 2025.

\end{thebibliography}
}

\clearpage
\setcounter{page}{1}
\maketitlesupplementary

\appendix
\setcounter{equation}{0}
\setcounter{figure}{0}
\setcounter{table}{0}

\section{Model Architecture Details}
\subsection{Position MLP Design} \label{app:positionmlp}
The Position MLP module is engineered to dynamically encode spatial information for variable-sized feature maps and subsequently transform these spatially-aware features through a high-capacity gated network. The architecture comprises two primary stages: dynamic position encoding extraction and a gated MLP transformation.

To accommodate varying input image resolutions without requiring interpolation of position embeddings, which can degrade performance, we employ a dynamic cropping strategy. We first initialize a large, learnable position embedding matrix, $\mathrm{P} \in \mathbb{R}^{\textit{N}*\textit{N}*D_{pos}}$, using a ``torch.nn.Embedding layer''. Here, $\textit{N} = 74$ is chosen to be larger than the maximum expected number of patches along any single dimension, and $D_{pos}$ is the dimension of the position embeddings. Given an input feature map from the vision encoder, $\mathrm{F}_{in} \in \mathbb{R}^{H^{'}*W^{'}*C_{in}}$, where $H^{'}$ and $W^{'}$ are the number of patches along the height and width respectively, and $C_{in} = 2048$. We extract a corresponding sub-matrix of position embeddings, $\mathrm{P}_{crop} \in \mathbb{R}^{H^{'}*W^{'}*C_{in}}$, by cropping the central $H^{'}*W^{'}$ region from the larger matrix $\mathrm{P}$. This ensures that the model always utilizes the most well-trained central portion of the positional space. The extracted position embeddings are then fused with the input features via element-wise addition:
\begin{equation}
    F_{fused} = F_{in} + P_{crop}
\end{equation}
This design provides flexibility in handling diverse aspect ratios and resolutions while maintaining a consistent and stable representation of spatial locality.

Following the spatial information fusion, the resulting tensor $F_{fused}$  is processed by a Gated Multi-Layer Perceptron (MLP). This structure allows the network to modulate the feature representation more effectively than a standard MLP. The input features (with dimension $C_{in} = 2048$) are first projected in parallel into a higher-dimensional space ($C_{hidden} = 16384$) by two separate linear layers, an up-projection layer ($W_{up}$) and a gating layer ($W_{gate}$). The output of the up-projection layer is used to multiplicatively gate the output of the gating layer, followed by an activation function ($\sigma$). The process can be formulated as:
\begin{equation}
    \mathrm{H} = \sigma(Linear_{gate}(\mathrm{F_{fused}})) \odot Linear_{up}(\mathrm{F_{fused}})
\end{equation}
where $\odot$ denotes element-wise multiplication. Subsequently, a final down-projection linear layer ($W_{down}$) maps the intermediate representation $\mathrm{H} \in \mathbb{R}^{H^{'}*W^{'}*C_{hidden}}$ back to the desired output dimension ($C_{out} = 2304$):
\begin{equation}
    \mathrm{F}_{out} = Linear_{down}(H)
\end{equation}
This gated mechanism empowers the model to control the information flow, enabling a more nuanced transformation of the spatially-enriched features.

\subsection{Simple ViT Design}
The Simple Vision Transformer (ViT) was developed as one of our initial exploratory methods for injecting VAE features into the Vision-Language Model (VLM). This architecture was conceived to investigate whether a dedicated, parameter-heavy module for processing these features could accelerate model training and improve convergence rates.
Unlike a traditional ViT that operates on image patches, our Simple ViT module directly processes latent features extracted from a VAE. The architecture bypasses any image-patching and patch-embedding stages. The operational flow is as follows:

\paragraph{Input Projection.} The input VAE features, denoted as $\mathrm{F}_{vae}$, are first passed through a linear projection layer. This layer maps the features from their original VAE latent dimension to the Simple ViT's internal working dimension of 2048, resulting in the projected feature map $\mathrm{F}_{proj}$.

\paragraph{Dynamic Position Encoding.} Before being processed by the Transformer layers, spatial context is explicitly added to $\mathrm{F}_{proj}$. To achieve this, we employ the identical dynamic center-cropping position encoding mechanism as detailed in Appendix~\ref{app:positionmlp}. A positional embedding, $\mathrm{P}_{crop}$, corresponding to the spatial dimensions of the feature map is extracted and added element-wise:
\begin{equation}
    \mathrm{F}_{pos} = \mathrm{F}_{proj} + \mathrm{P}_{crop}
\end{equation}

\paragraph{Transformer Layers.} The spatially-aware tensor, $\mathrm{F}_{pos}$, is then fed into a series of standard Transformer layers. The number of layers was a configurable hyperparameter in our experiments, with configurations such as two and six layers being tested, as referenced in the main body of the paper.

The primary hypothesis for this design was that a deeper, more expressive transformation of the VAE features through a dedicated Transformer stack would enable more effective integration with the broader VLM. However, while functionally sound, our empirical results showed that this approach was not the most efficient in terms of either computational cost or performance gains.

\subsection{Overall Architecture and Parameters}
\subsubsection{Our Architecture Design}
To ensure reproducibility and provide a clear reference, we detail the key hyper-parameter settings for the components of our WeMMU model in Table~\ref{tab:module_details}. This table covers the specific configurations for all modules, from the Vision Encoder and Large Language Model (LLM) to the Generation Expert and the newly introduced VAE branch. These parameters represent our choices after multiple rounds of experimental optimization, aimed at balancing model performance and computational efficiency. For a more conceptual description of how these components work together within our framework and the design philosophy, please refer to Section \textcolor{blue}{3.2} of the main paper.

\begin{table}[h]
    \centering
    \vspace{-1em}
    \caption{Hyper-parameter settings for WeMMU models}
    \vspace{-0.5em}
    \fontsize{8pt}{8pt}\selectfont
    \begin{tabular}{c|cc}
    \toprule
     ~ & ~  & \textbf{WeMMU}    \\
    \multirow{-2}{*}{\textbf{Module}}  & \multirow{-2}{*}{\textbf{Param.}}  & \textbf{settings}  \\
    \midrule
     \cellcolor{gray!5}~  & $N_{layer}$   & 32   \\
     \cellcolor{gray!5}~ & $d_{hidden}$ & 1280 \\
     \cellcolor{gray!5}\multirow{-3}{*}{\textbf{Vision Encoder(QWen2.5-VL)}} & \cellcolor{gray!10}$d_{out}$ & \cellcolor{gray!10}2048  \\
    \midrule
      \cellcolor{gray!5}~  & $N_{layer}$ & 36  \\
     \cellcolor{gray!5}\multirow{-2}{*}{\textbf{LLM (QWen2.5-VL)}} & \cellcolor{gray!10}$d_{hidden}$ &  \cellcolor{gray!10}2048  \\
    \midrule
      \cellcolor{gray!5}  ~  & $N_{layer}$ & 36 \\
      \cellcolor{gray!5}\multirow{-2}{*}{\textbf{Generation Expert}} & \cellcolor{gray!10}$d_{hidden}$  & \cellcolor{gray!10}2048   \\
    \midrule
      \cellcolor{gray!5}~ & $d_{in}$ & 32  \\
      \cellcolor{gray!5}\multirow{-2}{*}{\textbf{Linear Layer (after VAE Encoder)}} & \cellcolor{gray!10}$d_{out}$ & \cellcolor{gray!10}2048  \\
    \midrule
       \cellcolor{gray!5}~ & $d_{in}$ & 2048  \\
        \cellcolor{gray!5}~ & $d_{inter}$ & 16384  \\
        \cellcolor{gray!5}~ & $N_{layer}$ & 1  \\
      \cellcolor{gray!5}\multirow{-4}{*}{\textbf{PositionMLP}} & \cellcolor{gray!10}$d_{out}$ & \cellcolor{gray!10}2304  \\
    \midrule
       \cellcolor{gray!5}~ & $d_{in}$ & 3  \\
       \cellcolor{gray!5}~ & $d_{hidden}$ & 32  \\
      \cellcolor{gray!5}\multirow{-3}{*}{\textbf{VAE (Sana)}} & \cellcolor{gray!10}$f_{scale}$  & \cellcolor{gray!10}0.4107  \\
    \midrule
       \cellcolor{gray!5}~ & $N_{layer}$ & 20  \\
       \cellcolor{gray!5}~ & $d_{caption}$ & 2304  \\
       \cellcolor{gray!5}~ & $d_{in}$ & 32  \\
      \cellcolor{gray!5}\multirow{-4}{*}{\textbf{DiT (Sana)}} & \cellcolor{gray!10}$d_{out}$ & \cellcolor{gray!10}32  \\
    
    \bottomrule
    \end{tabular}
    \vspace{-2em}
    \label{tab:module_details}
\end{table}

\subsubsection{Comparative Analysis of Architectural Choices}
A central challenge in designing unified multimodal architectures is how to efficiently integrate understanding and generation capabilities. This challenge is particularly acute in scenarios involving a large number of input images, where the comparison between different architectural philosophies becomes most meaningful.

The first paradigm, represented by models like Bagel~\citep{bagel}, achieves powerful native image generation capabilities through end-to-end training within a single, unified model. The advantage of this approach lies in its ability to achieve a high degree of instruction following and semantic alignment. However, this comes at the cost of requiring massive, high-quality image-text datasets for training from scratch.

The second paradigm attempts to lower the training cost by "bridging" a pre-trained MLLM with a Diffusion Model, a strategy exemplified by works like MetaQueries~\citep{metaquery}. This approach cleverly leverages the powerful capabilities of existing models, allowing the MLLM to focus on understanding complex, multi-source contextual information. 

However, to compensate for the loss of fine-grained details during the MLLM's high-level processing, methods like Query-Kontext~\citep{querykontext} typically require injecting all the processed contextual features (e.g., dense image features, text embeddings) directly into the Diffusion Model. This is efficient, but as the number of source images increases, a critical bottleneck emerges. In these methods, the volume of contextual information grows dramatically. Injecting this complex and high-dimensional amalgamation of features directly into the Diffusion Model makes fine-tuning it exceptionally difficult. The generator must learn to dynamically disentangle and align features from an arbitrary number of inputs during the diffusion process, a highly unstable and complex optimization task.

In contrast, our WeMMU framework proposes a more advantageous "division of labor" strategy. In our design, the powerful MLLM undertakes the tasks it excels at: deeply understanding and reasoning about complex multimodal inputs, including parsing long text instructions and identifying relationships between multiple images. Crucially, the MLLM does not need to pass all raw or processed features directly to the generative model. Instead, through our proposed Noisy Query Tokens mechanism, it "summarizes" the complex generative intent and encodes it into a concise, robust distributed representation.

Intuitively, this clear division of labor not only preserves the potential for end-to-end optimization but also significantly reduces the "cognitive load" on both core components. We believe this presents a more efficient and promising path toward achieving sustainable and scalable unified multimodal generation frameworks.

\subsection{Training Details}
\paragraph{Optimizer and Learning Rate.} Across all four training stages, we utilized the AdamW optimizer with hyperparameters set to $\beta_1=0.9$, $\beta_2=0.95$, and a weight decay of $1.0e^{-5}$. The learning rate schedule for each stage included a specific number of warmup steps, followed by a cosine decay schedule, as detailed in Section \textcolor{blue}{4.1} of the main paper.

\section{Inference Strategy}
Our model employs distinct inference-time guidance strategies tailored to the specific task: text-to-image generation, single-image editing, and multi-image editing. These strategies leverage variations of Classifier-Free Guidance (CFG) to steer the diffusion process. It is important to note that in the following formulations, the noise prediction $\epsilon$ is generated by the diffusion model, which is conditioned on the latent representations produced by our Vision-Language Model (VLM) under different input scenarios.

\paragraph{Text-to-Image Generation.}
For standard text-to-image synthesis, we utilize a conventional CFG approach. Two conditioning latents are generated by the VLM: one from the target text prompt ($C_{con}$) and another from an empty or null text prompt ($C_{uncon}$). The diffusion model then predicts the corresponding noises, $\epsilon_{con}$ and $\epsilon_{uncon}$. The final noise prediction, $\epsilon_{pred}$, is a linear combination of these two predictions, formulated as:
\begin{equation}
    \epsilon_{pred} = \epsilon_{uncon} + \lambda(\epsilon_{con} - \epsilon_{uncon})
\end{equation}
Here, $\lambda$ is the guidance scale hyperparameter that controls the strength of the adherence to the text prompt. A higher value of $\lambda$ encourages the generation to more closely match the text description. In all text-to-image experiments, the value of $\lambda$ was set to 4.0.

\paragraph{Single-Image Editing.}  For single-image editing, a more sophisticated three-component guidance strategy is required to balance fidelity to the source image with the desired textual modification. We compute three distinct noise predictions: 
(1) $\epsilon_{uncon}$: Predicted from an unconditional latent, generated by the VLM with no image or text input.
(2) $\epsilon_{rec}$: Predicted from a reconstruction latent, where the VLM is conditioned on the source image and a simple reconstruction prompt (e.g., "Please reproduce this image"). 
(3) $\epsilon_{edit}$: Predicted from an editing latent, where the VLM is conditioned on the source image and the target editing prompt (e.g., "a cat wearing a top hat").
The final noise prediction is then calculated by combining these three components. This allows for separate control over reconstruction fidelity and edit strength:
\begin{equation}
    \epsilon_{pred} = \epsilon_{uncon} + \lambda_{rec}(\epsilon_{rec}-\epsilon_{uncon}) + \lambda_{edit}(\epsilon_{edit}-\epsilon_{rec})
\end{equation}
In this equation, $\lambda_{rec}$ is the reconstruction guidance scale, which encourages the model to preserve the structure and content of the original image. The term ($\epsilon_{edit}-\epsilon_{rec}$) represents the semantic direction of the edit itself. Consequently, $\lambda_{edit}$ controls the magnitude of the modification described in the editing prompt. This dual-guidance mechanism is crucial for achieving high-fidelity and semantically coherent image edits. In all Single-Image Editing experiments, the value of $\lambda_{rec}$ was set to 2.0 and the value of $\lambda_{edit}$ was set to 3.0.

\paragraph{Multi-Image Editing.} When performing edits that involve multiple source images (e.g., style transfer or subject swapping), the concept of a single "reconstruction" target becomes ill-defined. Therefore, we revert to a guidance strategy that is structurally similar to that of text-to-image generation but incorporates multi-image conditioning.
Specifically, we compute two noise predictions: (1) $\epsilon_{uncon}$: The standard unconditional noise prediction, as used in the other tasks. (2) $\epsilon_{multi}$: Predicted from a conditional latent, where the VLM is provided with the full set of source images and the target editing prompt.
The guidance is then applied as follows:
\begin{equation}
    \epsilon_{pred} = \epsilon_{uncon} + \lambda_{multi}(\epsilon_{multi} - \epsilon_{uncon})
\end{equation}
Here, $\lambda_{multi}$ is the guidance scale that determines how strongly the final output should reflect the combined context of the multiple input images and the textual instruction. In all Multi-Image Editing experiments, the value of $\lambda_{multi}$ was set to 3.0.

\section{Additional Generated Images}
\subsection{Text-to-Image Generation} \label{app:moreimages}
\begin{figure*}
    \centering
    \includegraphics[width=0.85\linewidth]{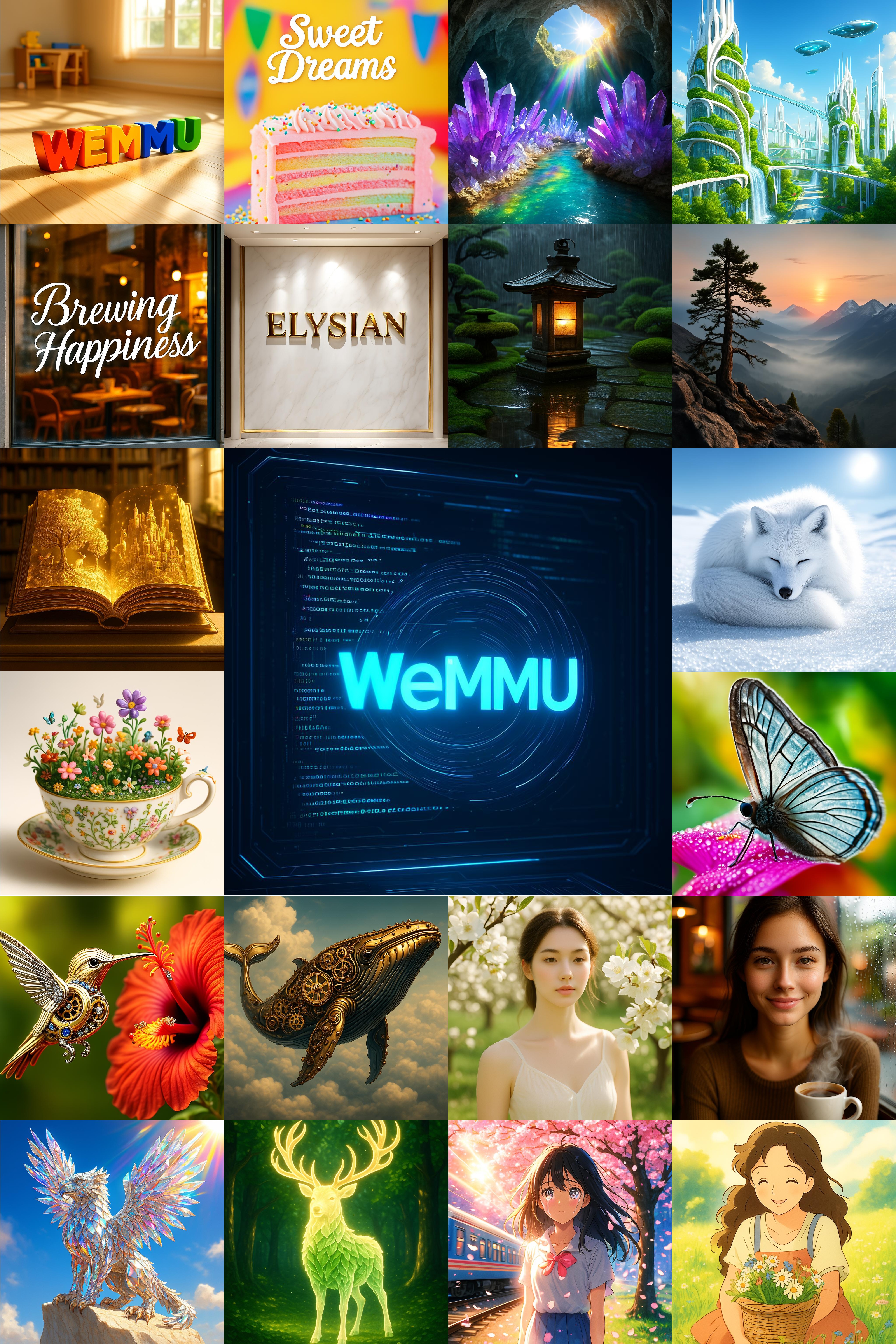}
    \caption{A gallery of diverse text-to-image generation results from our `WeMMU' model, synthesized at 1024x1024 resolution.}
    \label{fig:showed_t2i}
\end{figure*}

Fig.~\ref{fig:showed_t2i} showcases a diverse gallery of images synthesized by our `WeMMU' model. To provide full context for these high-fidelity results, the corresponding text prompts for each image are detailed below, following a top-to-bottom, left-to-right reading order.

\begin{description}
    \item[1] \textit{The model name `WEMMU' spelled out with colorful, polished wooden toy blocks on the light wood floor of a child's playroom. The room is flooded with warm, bright sunlight from a large window, creating soft, long shadows. Shallow depth of field.}
    
    \item[2] \textit{A beautiful macro photograph of a slice of pastel rainbow layer cake. On the top, the words `Sweet Dreams' are written in elegant, creamy white icing, adorned with tiny, colorful sugar sprinkles. The background is a bright, cheerful, out-of-focus party scene.}
    
    \item[3] \textit{A vast, breathtaking crystal cave, where the ceiling is open to the bright sky above. Pure, brilliant sunlight streams down, striking giant, perfectly formed crystals of amethyst and quartz, refracting into a thousand vibrant rainbows that illuminate the entire cavern. A crystal-clear river flows gently through the cave floor.}
    
    \item[4] \textit{A panoramic, sun-drenched vista of a Solarpunk utopian city. Gleaming white towers with organic, flowing lines are covered in lush vertical gardens and waterfalls. Shimmering, elegant glass sky-bridges connect the buildings, and flying, bio-luminescent vehicles drift through the clean air. The scene is bright, optimistic, and incredibly detailed.}
    
    \item[5] \textit{The phrase `Brewing Happiness' is written in beautiful, elegant white script on the clean glass window of a charming coffee shop. Through the window, the cafe's warm, brightly lit, cozy interior is visible.}
    
    \item[6] \textit{The word `Elysian' embossed in shimmering gold foil on a pristine, polished white marble wall at the entrance to a luxury boutique. Soft, bright, diffused lighting from above creates gentle highlights and shadows on the letters.}

    \item[7] \textit{A single, ancient stone lantern in a serene Japanese garden during a heavy downpour. Its warm, gentle light creates a small haven of peace, and its reflection shimmers beautifully on the wet, mossy stone path.}
    
    \item[8] \textit{A wide-angle landscape shot from a mountain peak. In the foreground, a gnarled, ancient pine tree clings to the rocks. In the mid-ground, a misty valley unfolds. In the far distance, the sun rises behind another range of snow-capped mountains.}
    
    \item[9] \textit{An open, magical book lies on a pedestal in a sunlit library. The pages are made of hammered gold leaf, and as the pages turn, intricate 3D models of trees, animals, and cities emerge from the pages in shimmering, golden light.}
    
    \item[10] \textit{A futuristic, semi-transparent holographic interface displaying glowing blue lines of code and, in the center, the word `WeMMU' rotates slowly in 3D space.}
    
    \item[11] \textit{A flawless, pure white arctic fox curled up on pristine, sparkling snow under a bright, clear arctic sun. The brilliant light reveals the incredible detail and texture of its thick fur against the crystalline snow. White-on-white photography, every strand visible.}
    
    \item[12] \textit{A whimsical, miniature, exquisitely detailed garden of blooming flowers and tiny, magical creatures, all contained within an elegant porcelain teacup. The scene is shot with a macro lens under bright, soft studio lighting.}
    
    \item[13] \textit{An extreme macro photograph of a Glasswing butterfly (Greta oto) resting on a vibrant, dew-covered orchid. The bright morning light illuminates its stunningly transparent wings, highlighting the delicate, iridescent borders. Every tiny detail is in perfect focus.}
    
    \item[14] \textit{A beautiful clockwork hummingbird, crafted from polished gold, gleaming silver, and tiny, sparkling sapphires. It hovers beside a vibrant, sun-drenched hibiscus flower, its intricate gears and mechanisms fully visible and crisply rendered. Macro photography, bright natural light.}
    
    \item[15] \textit{A majestic whale gracefully swimming through the clouds, its body a stunning fusion of organic flesh and intricate, polished brass clockwork mechanisms. Steampunk, fantasy.}
    
    \item[16] \textit{A serene and beautiful woman with delicate features, standing in an orchard of blooming white apple blossoms. Bright, soft sunlight filters through the branches, highlighting the translucent quality of the petals and the soft texture of her skin.}

\begin{figure*}
    \centering
    \includegraphics[width=0.88\linewidth]{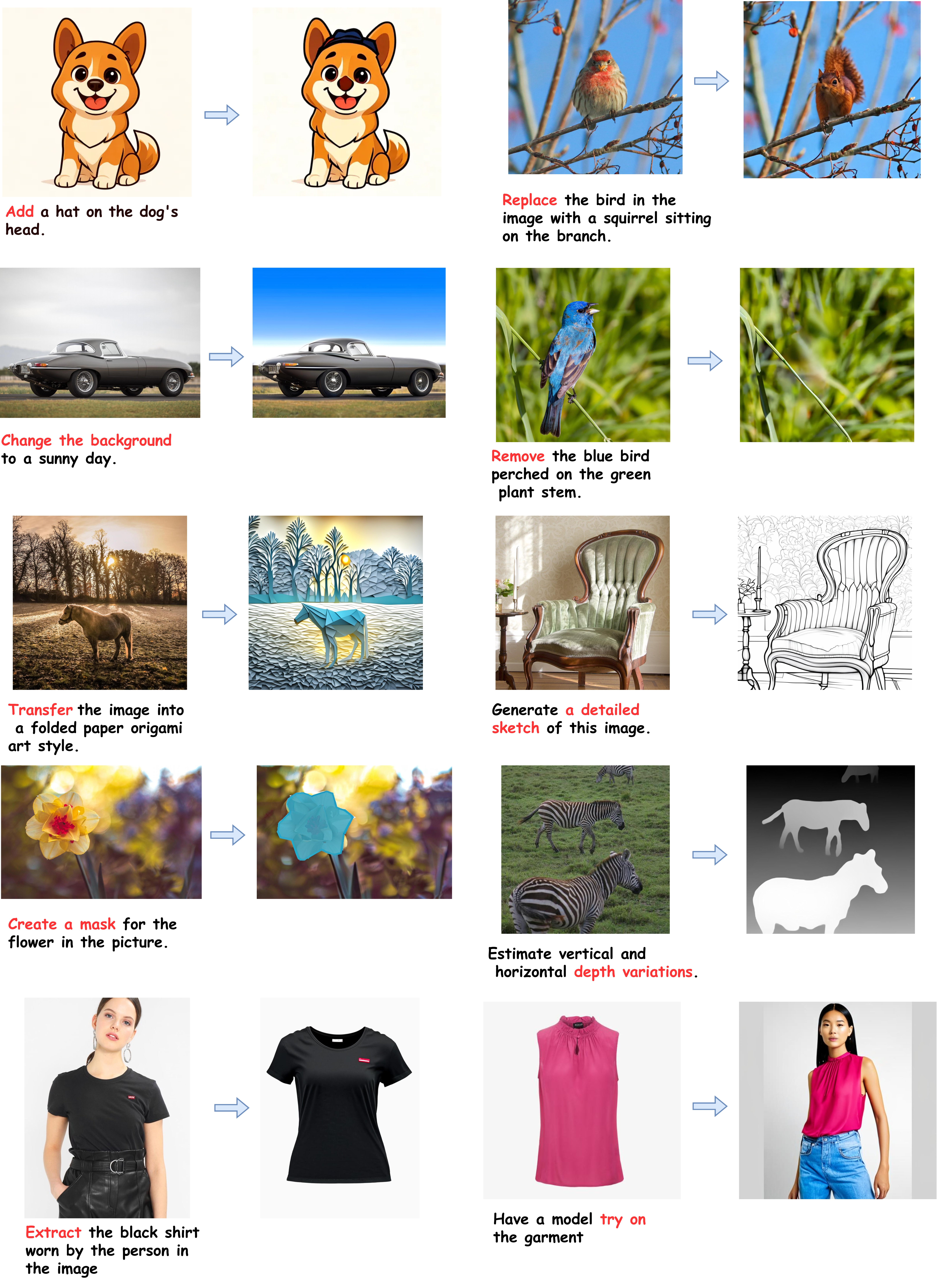}
    \caption{Illustrative examples of single-image editing performed by our `WeMMU' model. Each pair shows the original image (left) and the edited result (right) based on the annotated prompt.}
    \label{fig:showed_i2i}
\end{figure*}

\begin{figure*}
    \centering
    \includegraphics[width=0.80\linewidth]{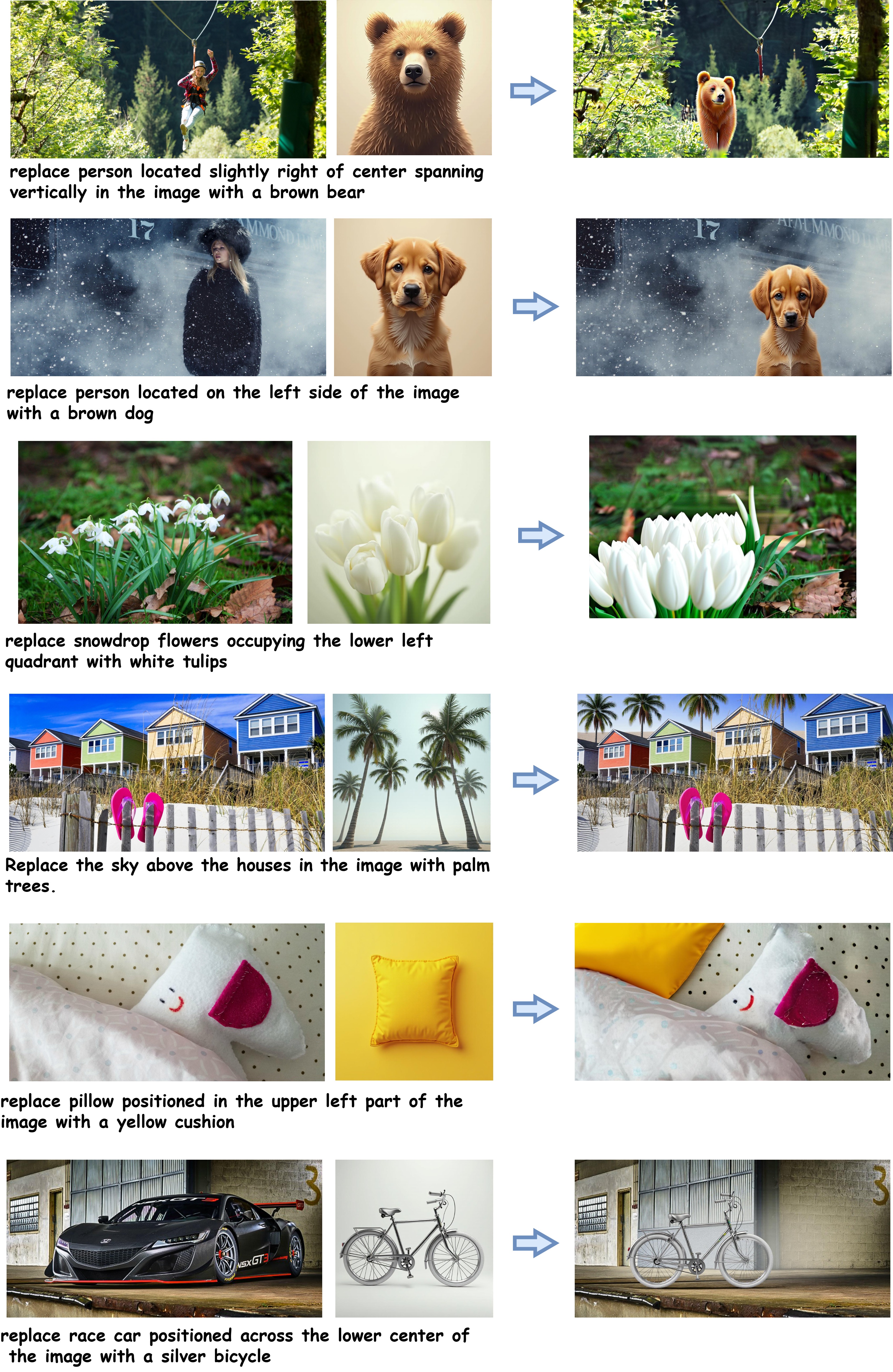}
    \caption{Illustrative examples of multi-image editing. Each set displays the two source images alongside the final composite image generated by `WeMMU' based on the annotated prompt.}
    \label{fig:showed_mul_i2i}
\end{figure*}
    
    \item[17] \textit{Ultra-realistic portrait of a beautiful young woman with a gentle smile, sitting inside a cozy, brightly lit cafe. Soft natural light streams through a large window covered in fresh raindrops. The warm interior glow reflects in her eyes, and you can see the delicate texture of her skin and the steam rising from her coffee cup.}
    
    \item[18] \textit{A majestic griffin perched on a sun-drenched marble cliff. Its magnificent wings are not made of feathers, but of intricate, interlocking pieces of opalescent crystal that refract the bright sunlight into a dazzling spectacle of color.}
    
    \item[19] \textit{A majestic and benevolent forest spirit, resembling a great stag, whose body appears to be woven from pure, solid light and living, green leaves. Its grand antlers glow with a warm, gentle light. It stands peacefully in the center of a sun-dappled, magical forest, radiating an aura of calm and life.}
    
    \item[20] \textit{A vibrant, emotional anime scene in the style of Makoto Shinkai. A beautiful young woman with sparkling eyes stands under a blooming cherry blossom tree as a train passes by. The wind sweeps petals around her. The scene is characterized by dramatic lens flare, incredibly detailed backgrounds, and a bright, saturated color palette.}
    
    \item[21] \textit{A beautiful, gentle young woman with a warm smile, in the charming, hand-drawn aesthetic of Studio Ghibli. She is sitting in a lush, sunlit meadow, weaving a crown from a basket of fresh wildflowers. The scene is peaceful, nostalgic, and filled with warm, natural light.}
    
\end{description}

\subsection{Single-Image Editing}
Beyond image generation, our `WeMMU' model also supports a variety of single-image editing tasks. Fig.~\ref{fig:showed_i2i} showcases this versatility across a range of creative and functional edits, with prompts annotated directly on the images. The demonstrated capabilities include fundamental semantic changes (adding, removing, replacing), stylistic transfers (origami, sketch), and utility-oriented functions such as depth map generation and virtual try-on, highlighting the model's ability to respond to a diverse set of user instructions.

\subsection{Multi-Image Editing}
Finally, we explore the model's capabilities in multi-image editing, a task that requires synthesizing a new image based on two source images and a guiding text prompt. Fig.~\ref{fig:showed_mul_i2i} presents examples of this functionality. While the results demonstrate that `WeMMU' is capable of following these compositional instructions—successfully identifying, extracting, and arranging specified features from both source images into a new target image—we also observe certain artifacts in the final outputs. Some generated images may exhibit a noticeable ``pasted-on'' appearance at the seams of integrated elements, occasionally accompanied by a loss of fine-grained detail. We attribute these limitations primarily to the scarcity and inconsistent quality of the multi-image conditioning data used during training.

Nevertheless, these examples validate that the core mechanism for multi-image feature extraction and compositional synthesis is functional. This establishes a promising foundation for future work, where performance can be substantially improved with access to larger, higher-quality datasets.

\end{document}